\documentclass[runningheads]{llncs}

\usepackage{eccv}

\usepackage{eccvabbrv}

\usepackage{graphicx}
\usepackage{booktabs}
\usepackage{tcolorbox}
\usepackage[pagebackref,breaklinks,colorlinks]{hyperref}
\usepackage[table]{xcolor}
\usepackage{algorithm}
\usepackage{algpseudocode}
\tcbuselibrary{listings,breakable}
\usepackage{tcolorbox}
\usepackage{multirow}
\usepackage{booktabs}   %
\usepackage{multirow}   %
\usepackage[table]{xcolor} %
\usepackage{amssymb}    %
\usepackage[table]{xcolor}
\usepackage{colortbl}

\usepackage[table]{xcolor}
\usepackage{tikz}
\usetikzlibrary{tikzmark,calc}

\usepackage{wrapfig}
\usepackage{amssymb}
\usepackage{pifont}

\definecolor{verylightblue}{RGB}{220, 235, 255}

\usepackage[accsupp]{axessibility}  %

\usepackage{hyperref}

\usepackage{orcidlink}

\begin{document}

\title{TTA-Vid: Generalized Test-Time Adaptation for Video Reasoning} 

\titlerunning{TTA-Vid}

\author{Soumya Shamarao Jahagirdar$^1$\thanks{Equal contribution.}\quad Edson Araujo$^{1*}$\quad Anna Kukleva$^6$ \\ 
M. Jehanzeb Mirza$^2$ \quad Saurabhchand Bhati$^2$\quad Samuel Thomas$^3$ \\ Brian Kingsbury$^3$ \quad Rogerio Feris$^{3,4}$\quad
James R. Glass$^2$\quad Hilde Kuehne$^{4,5}$\\}
\institute{\small
$^1$ University of Tübingen,
$^2$ MIT,
$^3$ IBM Research,
$^4$ MIT-IBM Watson AI Lab, \\
\small
$^5$ Tuebingen AI Center, \quad
$^6$ Max Planck Institute for Informatics, SIC \\
\email{\{soumya-shamarao.jahagirdar, edson.roteia-araujo-junior\}@uni-tuebingen.de} \\
}

\authorrunning{S. Jahagirdar et al.}

\maketitle
\begin{abstract}
  Recent video reasoning models have shown strong results on temporal and multimodal understanding, yet they depend on large-scale supervised data and multi-stage training pipelines, making them costly to train and difficult to adapt to new domains.
  In this work, we leverage the paradigm of Test-Time Reinforcement Learning on video-language data to allow for adapting a pretrained model to incoming video samples at test-time without explicit labels.
  The proposed test-time adaptation for video approach (TTA-Vid) 
  performs step-by-step reasoning at inference time on multiple frame subsets.
  We then use a batch-aware frequency-based reward computed across different frame subsets as pseudo ground truth to update the model.
  It shows that the resulting model trained on a single batch or even a single sample from a dataset, is able to generalize at test-time to the whole dataset and even across datasets. 
  Because the adaptation occurs entirely at test time, our method requires no ground-truth annotations or dedicated training splits. 
  Additionally, we propose a multi-armed bandit strategy for adaptive frame selection that learns to prioritize informative frames, guided by the same reward formulation.  
  Our evaluation shows that TTA-Vid yields consistent improvements across various video reasoning tasks and is able to outperform current state-of-the-art methods trained on large-scale data. 
  This highlights the potential of test-time reinforcement learning for temporal multimodal understanding. \footnote{All code, data, and checkpoints will be made available.}
\end{abstract}
    
\section{Introduction}
\label{sec:intro}

Understanding the content of long videos and reasoning over it remains a fundamental challenge in video comprehension and multimodal learning. Recent progress in large Vision Language Models (VLMs) such as the InternVL \cite{chen2024internvl, zhu2025internvl3, wang2025internvl3_5} or QwenVL \cite{qwen_vl, bai2025qwen25vl} series, and video reasoning models such as Video-R1 \cite{feng2025video}, VideoRTS \cite{wang2025video}, Video-RFT \cite{wang2025videorft}, VideoChat-R1 \cite{li2025videochatr1}, VideoChat-R1.5 \cite{yan2025videochatr1.5}, and Video-R2 \cite{maaz2025videor2} has brought remarkable advances in e.g. captioning or question answering tasks. However, when applied to lengthy, highly structured, and conceptually rich videos, these generic MLLMs struggle to produce coherent reasoning and accurate answers. Instructional videos have become particularly important in evaluating MLLM performance in this domain as they often feature visual and verbal content that is inherently structured over time, with one information building up on another. Building models that can capture such content provides a compelling test of whether AI systems can capture relevant information over time and reason and generalize over it.
Despite their potential, existing video reasoning models face critical limitations in this context, as they typically adopt multi-stage pipelines that involve supervised finetuning followed by reinforcement learning or reward model optimization. These procedures are not only computationally intensive, especially in the case of video, but can also limit the model's ability to adapt to new domains without retraining. Second, as those models need a significant amount of training data, they leverage large-scale datasets such as ActivityNet-QA \cite{yu2019activitynet}, which often contain shorter clips. As a result, the models struggle to capture longer, more complex reasoning structures, as can e.g. be found in real-world content.

\begin{figure}[t]
    \vspace{-0.5\baselineskip} %
    \centering
    \includegraphics[width=\linewidth]{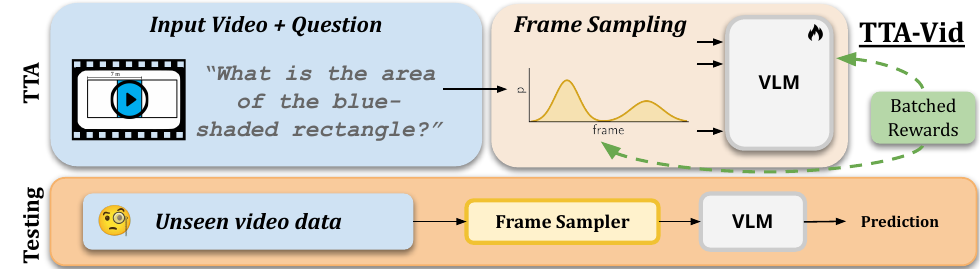}
    \caption{TTA-Vid adapts vision-language models at test-time by sampling multiple frame subsets, enforcing majority-consistency among generated answers, and updating a frame-importance distribution via a multi-armed bandit. This allows a test-time adaptation without labels while selecting frames most relevant for reasoning.}
    \label{fig:teaser}
    \vspace{-8mm} %
\end{figure}

To address those limitations, we propose a test-time adaptation framework for video reasoning. Unlike few-shot learning, which relies on a small labeled support set, test-time adaptation operates entirely on unlabeled data without any annotations. Specifically, given an instructional video and a question, the proposed method samples multiple sets of frames and generates multiple candidate reasoning traces and answers. A batch-wide frequency reward is computed by measuring the empirical probability of each answer across all generated outputs, combined with an entropy-based confidence penalty, allowing the model to reinforce frequently occurring answers while suppressing high-uncertainty generations. While related test-time RL methods such as TTRL \cite{zuo2025ttrl} and TTRV \cite{singh2025ttrv} target language models and single-image tasks, respectively, TTA-Vid exploits a fundamental property of video: videos contain far more frames than a model can process at once. Sampling different frame subsets naturally produces diverse rollouts grounded in different visual evidence, and the resulting reward signal can be leveraged to learn which frames contribute most. 
Beyond that, we found that models adapted on a single batch generalize to the full dataset and even across datasets. Thus, training on a single batch of 32 unlabeled samples is sufficient to improve performance on the entire unseen test set, and the gains even transfer across datasets and even improves over a full test-time training on every batch. This indicates that the model acquires transferable reasoning capabilities rather than overfitting to dataset-specific patterns. This is especially relevant for the case of video, as test-time RL is still time intensive. To our best knowledge, we are the first to report this phenomenon for test-time video reasoning adaptation.   

Finally, we introduce a frame distribution learning mechanism based on the multi-armed bandit formulation. Instead of relying on fixed frame sampling, our method maintains a learnable distribution that assigns importance scores to frames based on their contribution to successful reasoning. This distribution is optimized using test-time reinforcement learning signals derived from a batch-wide frequency reward that combines empirical answer frequencies with an entropy-based confidence penalty, gradually learning which frames matter most for answering each question. As test-time training progresses, the model converges toward a video-specific sampling policy, highlighting frames that are semantically important. This frame selection process benefits the final systems in two ways. First, the fact that the model can focus on better frames during the training can help to generate better rollouts and thus more traces relevant for the reward. Second, it further shows that the learned distribution also generalizes beyond a single test batch and improves over vanilla based lines such as CLIP-based matching. 
The resulting approach requires no additional supervised data, but instead leverages the inherent redundancy in long videos to build a self-supervised adaptation loop at test time.

We evaluate the proposed method on five benchmarks, VideoMMMU \cite{hu2025videommmu}, MMVU \cite{zhao2025mmvu}, SciVideoBench \cite{deng2025scivideobench}, VideoMME \cite{fu2025videomme}, and LongVideoBench \cite{wu2024longvideobench}, and two distinct vision-language backbones, InternVL-3 \cite{zhu2025internvl3} and Qwen2.5-VL \cite{bai2025qwen25vl}.
The evaluation shows that the proposed test-time adaptation approach consistently improves answer consistency and is able to outperform current methods trained on large-scale ground-truth data based on only a single batch of 32 test samples without any labels. It further shows that the improvement is not only limited to the respective test dataset, but also extends to unseen test data, allowing for a generalized test-time adaptation. 
Our evaluation overall shows that test-time reinforcement learning can serve as a powerful and efficient tool for video reasoning, paving the way for better models and methods that can autonomously adapt and reason over long, structured video content.

The contributions of this work can be summarized as follows:
(a) TTA-Vid: A test-time reinforcement learning framework that adapts vision-language models to video reasoning tasks without any labeled data, curated training splits, or pre-generated reasoning traces, yet outperforms video reasoning models trained on large-scale supervised datasets.
(b) A multi-armed bandit strategy that learns frame importance distributions for efficient and interpretable reasoning using the test-time training reward signals.
(c) An extensive evaluation that focuses on the case of generalized test-time adaptation across multiple instructional video question-answering datasets and model backbones, showing state-of-the-art performance even compared to models trained on large-scale data and validating the effectiveness of our approach.

\section{Related Work}
\label{sec:related_works}

\subsection{Video Reasoning Models}

The success of large language models (LLMs) \cite{brown2020language,llama_3_2024, dataset_Analyst_Subpopulation_Structure_eccv_2024,Training_language_models_to_follow_instructions_nips_2022,llama_2023,vicuna_2023,Language_models_are_unsupervised_multitask_learners} has motivated the extension of their capabilities to multimodal tasks. This lead to the emergence of vision-language models allowing models to consider visual content as well as video models which interpret dynamic visual content \cite{language_is_not_all_you_need_2023, Palme_2023, visual_intruction_tuning_2023, qwen_vl, blip_2_icml_2023, li2022blip,zhu2025internvl3,bai2025qwen25vl,an2025llava15onevision,zhang2024video,li2024llavaov,wang2024internvideo2,bai2025qwen2,zhang2024simplellovi}. However, most models like LLaMA-VID\cite{li2024llamavid}, VideoLLaMA2 \cite{cheng2024videollama2}, LongVA \cite{zhang2024longva}, VISA \cite{yan2024visa} among others focus on video perception tasks.
On the other hand, works inspired by reasoning in language models \cite{guo2025deepseek,jaech2024openai,shao2024deepseekmath,liu2025understandingdrgrpo,yu2025dapo,team2025kimi} such as \cite{xu2025llavacot,thawakar2025llamav-o1,deng2025openvlthinker,yang2025r1onevision,chen2025vlaa-thinker,shen2025vlm-r1} target image-based reasoning using hand-crafted CoT structures. Several recent works, \cite{meng2025videocap,wang2025timer1,sun2025videosalmon,zhang2025tinyllava,dang2025reinforcing,wang2025videorft} have extended vision-language reasoning to the video domain. Video-R1 \cite{feng2025video} introduces a T-GRPO algorithm, specifically designed to handle temporal information in videos. It utilizes two datasets: Video-R1-CoT-165k for supervised finetuning (SFT) and Video-R1-260k for reinforcement learning (RL) training. Video-RFT \cite{wang2025videorft} proposes a multi-expert driven, cognition-inspired CoT curation pipeline. In this framework, an LLM first generates preliminary CoTs based on rich, structured, and literal representations of video content. A VLM then refines these CoTs by conditioning them on the actual video input. This process results in two datasets: VideoRFT-CoT-102K for SFT and VideoRFT-RL-310K for RL training. In contrast, Video-RTS \cite{wang2025videorts} presents a different approach by combining efficient RL with a video-adaptive test-time scaling (TTS) strategy. All these methods depend on ground truth annotations and large, high-quality CoT and RL data. Compared to that, test-time reinforcement learning approach leverages the majority answer as a reward signal allowing a training without annotation. %

\subsection{Test-Time Training and Adaptation}
Various LLM and VLM approaches have explored leveraging unlabeled data through test-time adaptation and unsupervised learning. Parameters of the models are adjusted at inference or the methods learn from external unlabeled datasets by optimizing objectives such as RL rewards, entropy minimization, auxiliary self-supervised loss among others \cite{sun2019test,sun2024learning,behrouz2024titans,akyurek2024surprising,zuo2025ttrl,sun2020test,sun2024learning,mirza2023actmad}. TTRL \cite{zuo2025ttrl} utilizes repeated sampling strategy during the rollout phase to accurately estimate the labels, followed by a majority voting reward applied on the given unlabeled data, and TTRV \cite{singh2025ttrv} extends it by combining the frequency-based rewards with entropy regularization on vision tasks such as classification and VQA.
Complementary to these, recent work has shown that reinforcement learning can generalize from remarkably few examples. 1-shot RLVR \cite{wang2025rlvrwithoneexample} demonstrates that RL with verifiable rewards on a single training example is sufficient to substantially improve mathematical reasoning in LLMs, while LIMR \cite{li2025limr} shows that a strategically selected subset of training samples can match or outperform full-scale RL datasets, suggesting that sample selection matters more than data scale. These findings motivate our extension to the video domain, where labeled data is particularly scarce.
Building on these ideas, our method extends test-time adaptation to video understanding tasks, specifically for educational videos containing reasoning based questions. We incorporate an adaptive frame sampling strategy based on a multi-armed bandit problem, which complements the test-time adaptation, resulting in a novel framework for video reasoning on educational and lecture video content.

\begin{figure*}[t]
	\centering
	\includegraphics[width=\textwidth]{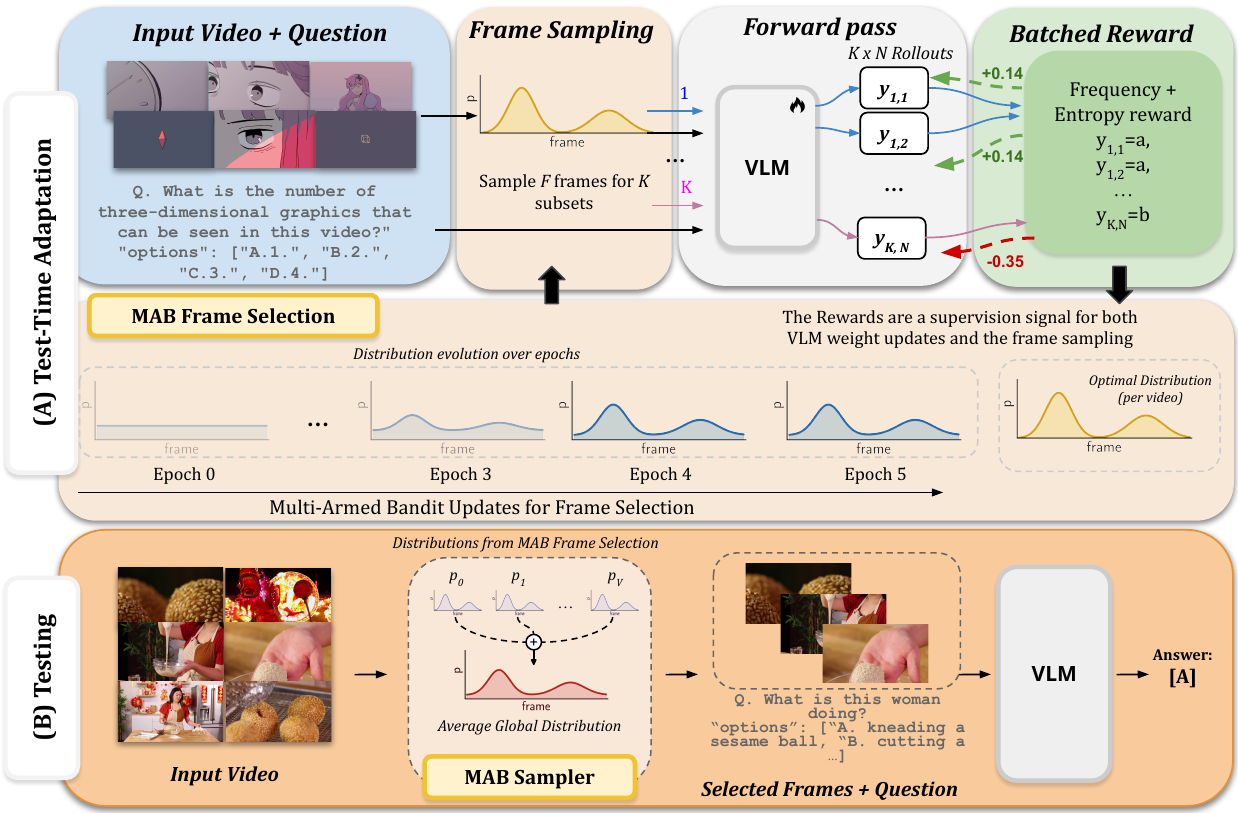}
	\caption{Overview of TTA-Vid: Our method performs test-time adaptation of model parameters through a batch-aware reinforcement learning objective and adaptively selects the most informative frames using a multi-armed bandit approach. Both components leverage a shared reward signal computed across multiple video frame subsets, enabling the model to jointly learn what to predict and which frames to attend to.}
	\label{fig:method_overview}
    \vspace{-5mm}
\end{figure*}

\section{TTA-Vid}

Our method proposes a framework that adapts a pretrained model at test time without ground-truth labels. While the adaptation is performed on a small batch of unlabeled samples, the resulting improvements generalize beyond the training batch to unseen test data and even transfer across datasets.

It is composed of two components that work simultaneously: (i) The \textbf{Test-Time Adaptation (TTA)} procedure updates the model parameters using a reinforcement learning objective guided by a batch-aware reward signal. Instead of relying on a single view of a video, we sample multiple subsets (or ``views'') of frames from the same video and generate multiple candidate outputs for each subset. Aggregating these outputs allows us to estimate a batch-level answer distribution that provides a stable reward signal, encouraging the model to produce consistent predictions across different frame subsets. 
(ii) We leverage an \textbf{Adaptive Frame Selection}, formulated as a multi-armed bandit problem. During adaptation, we sample different frame subsets based on a sampling distribution over frames. We use the reward signal formulated for the TTA component to update the sampling distribution and to give more weight to frames from successful rollouts. 
After each step, these distributions are updated to guide frame sampling for the next step. %
Figure~\ref{fig:method_overview} provides an overview of the approach. 
In the following, we first describe the test-time adaptation procedure and batch reward formulation, followed by the inference protocol using the adapted model.

\subsection{Test-Time Adaptation with Batch-Wide Frequency Reward}
\label{sec:tta}

We begin by representing an input video $\mathcal{V}$ as an ordered sequence of frames: $\mathcal{V} = (f_1, f_2, \dots, f_T)$. From this video, we sample $K$ subsets $\{S_1, S_2, \dots, S_K\}$ with $F$ frames $S_k=(\hat{f_{t_1}}, \dots, \hat{f_{t_F}}) \in V$ with $(t_1, \dots, t_F) \in T$. Given a prompt $x$ as the combination of a question and the frame subset $S_k$, the model, parameterized by $\theta$, generates an output $y_k$ from its policy $\pi_{\theta}(y_k|x, S_k)$. Note that practically, we pool a new set of subsets at each epoch, following the learned distribution. For simplicity of notation, we assume the notation to refer to a single epoch. 

To construct the reward signal, for each of the $K$ subsets we generate $N$ rollouts resulting in $N$ outputs, creating a total pool of $K \times N$ output candidates, with a single candidate denoted as $\{y_{k,n}\}$ for $k=1,\dots,K$ and $n=1,\dots,N$, and all samples of the pool relating to prompt $x$, thus the same question and the same video, and corresponding to $n$ rollouts for a single subset $k$.

Our reward formulation extends the frequency-based reward concept from TTRV \cite{singh2025ttrv} which estimates the empirical probability of an answer to be correct by calculating frequency within the answers extracted from the rollouts for a single image. Compared to that, we estimate the empirical probability of an answer based on all $N$ rollouts across all $K$ subsets. This approach leverages the diversity of views of a video to create a more stable and robust reward signal.
We use a fixed answer extractor function, $h$ (e.g., a regular expression), to parse the specific answer from each generated output string $y_{k,n}$. Let $\mathcal{A}$ be the set of all unique answers. We define the counts and the empirical frequency for each unique answer $a \in \mathcal{A}$ as:
\begin{equation}
    c(a) = \sum_{k=1}^{K} \sum_{n=1}^{N} \mathbf{1}\big[h(y_{k,n}) = a\big], \quad
    p(a) = \frac{c(a)}{\sum_{a' \in \mathcal{A}} c(a')}
\end{equation}
To promote convergence and control for diversity \cite{singh2025ttrv}, we incorporate an entropy-based confidence weight. We calculate the normalized entropy of the answer distribution $p$:
\begin{equation}
\begin{aligned}
H(p) &= -\sum_{a\in\mathcal{A}} p(a)\log p(a), \\
H_{\text{norm}}(p) &= \frac{H(p)}{\log |\mathcal{A}|} \in [0,1].
\end{aligned}
\end{equation}
The final reward for an individual output $y_{k,n}$ combines the frequency score with this entropy penalty:
\begin{equation}
r(y_{k,n}) \;=\; p\big(h(y_{k,n})\big) - \alpha \cdot H_{\text{norm}}(p),
\label{eq:reward_final}
\end{equation}
where $\alpha$ is a hyperparameter that controls the strength of the entropy penalty. This reward structure encourages outputs that are frequent across the batch while simultaneously penalizing the model for high uncertainty (high entropy) in its overall answer distribution, pushing it toward more confident predictions.

The reinforcement learning objective is to maximize the expected reward, and the model parameters $\theta$ are updated accordingly:
\begin{equation}
\theta \leftarrow \theta + \eta \,
\nabla_{\theta} \, \mathbb{E}_{S_k} \,
\mathbb{E}_{y \sim \pi_{\theta}(\cdot|S_k)}
\big[\, r(y) \,\big],
\end{equation}
where $\eta$ is the learning rate. Following \cite{zuo2025ttrl}, we implement this using Group Relative Policy Optimization (GRPO)~\cite{shao2024deepseekmath}.

\subsection{Multi-Armed Bandit Adaptive Frame Selection}
\label{sec:Adaptive_Frame_Selection}
While the TTA process adapts the model's parameters, we further incorporate an adaptive frame selection that learns an optimal policy for sampling video subsets. We treat each of the $T$ frames as an ``arm" in a contextual multi-armed bandit problem \cite{auer2002nonstochastic}, where the goal is to learn a probability distribution that favors frames that are most informative for the given task.

We maintain nonnegative \emph{weights} $\mathbf{w} = [w_1, w_2, \dots, w_T]$ for the $T$ frames, initializing $w_t = c$ for all $t$, where $c$ is the normalized CLIP score obtained based on the similarity for the $T$ frames and the question. From these, we define a learnable probability distribution over frames as $\mathbf{p} = [p_1, p_2, \dots, p_T]$, where
\begin{equation}
p_t = \frac{w_t}{\sum_{j=1}^T w_j}.
\label{equation:p_definition}
\end{equation}
At each epoch, the indices $(t_1, \dots, t_F) \in T$ for the frames of each subset $S_k$ are sampled stochastically according to $\mathbf{p}$.

The reward signal calculated for the TTA component (see Eq.~\ref{eq:reward_final}) is repurposed to guide the frame selection. For each subset $S_k$, we calculate its average reward by averaging the rewards of the $N$ outputs generated from it:
\begin{equation}
\bar{r}_k = \frac{1}{N} \sum_{n=1}^{N} r(y_{k,n}).
\end{equation}
This score, $\bar{r}_k$, reflects how informative the frames in subset $S_k$ were in contributing to high-frequency, high-confidence answers.
To update the frame distribution, we use a multiplicative weights algorithm. We first establish a baseline reward, $\bar{r}_{\text{baseline}} = \frac{1}{K} \sum_{k=1}^{K} \bar{r}_k$, which represents the average performance across all subsets. The probabilities of frames in subsets that performed better than this baseline are increased, while those in underperforming subsets are decreased.

The update for the weight $w_t$ of each frame $t$ is given by:
\begin{equation}
w_t^{\text{new}} = w_t \cdot \exp\left( \eta_{fs} \sum_{k=1}^K (\bar{r}_k - \bar{r}_{\text{baseline}}) \cdot \mathbf{1}[t \in S_k] \right),
\label{eq:mab}
\end{equation}
where $\eta_{fs}$ is the frame selection learning rate and $\mathbf{1}[t \in S_k]$ is an indicator function. We then form the new sampling probabilities by normalizing:
\begin{equation}
p_t^{\text{new}} = \frac{w_t^{\text{new}}}{\sum_{j=1}^T w_j^{\text{new}}}.
\end{equation}

\subsection{Generalization of the Adapted Model} 
After the test-time adaptation stage, we evaluate the adapted vision-language model on the full test dataset. 
During adaptation, the model parameters $\theta$ are optimized together with the individual probability distribution for a set of videos $(\mathcal{V}_1, \dots, \mathcal{V}_V )$ resulting in a set of probability distribution $\mathbf{p}_1 \dots \mathbf{p}_V$, with each training video yielding a learned frame distribution $\mathbf{p}_i$. 
To obtain a global frame distribution, we average these final distributions across the samples used for test-time adaptation $\bar{p}=\frac{1}{V}\sum_{i=1}^{V} p_i$. 
$V$ denotes the number of videos used during adaptation. 
The resulting distribution $\bar{p}$ represents a global frame prior that reflects frames that were consistently identified as informative across the adaptation set. 
For each test video, adapted model is used to predict the final answer with frames sampled according to $\bar{p}$.

\section{Experiments}

\definecolor{lightblue}{HTML}{D9EAF7}
\definecolor{lightyellow}{HTML}{FFF9C4}

\begin{table*}[t]
    \centering
    \resizebox{1.0\textwidth}{!}{%
    \begin{tabular}{@{}lcccccccc@{}}
    \toprule
    {Model} & {\#params} & {\#frames} & {VideoMMMU} & {MMVU} & {SciVideoBench} & {VideoMME} & {LongVideoBench} & {Avg.} \\
    \midrule
    Random & - & - & $14.00$ & $20.00$ & $10.00$ & $25.00$ & $20.00$ & $17.80$\\ \midrule
    \textit{Proprietary Models} & & & & & & \\ \midrule
    
    GPT-4o \cite{hurst2024gpt} & $-$ & $-$ & $61.22$ & $75.40$ & $24.90$ & $71.90$ & $66.70$ & $60.02$ \\
    Gemini 1.5 Flash \cite{team2024gemini} & $-$ & $-$ & $49.78$ & $58.80$ & $-$ & $70.30$ & $61.60$ & $-$\\\midrule
    \textit{Non-Reasoning Models} & & & & & & \\ \midrule
    LLaVA-OneVision \cite{li2024llavaov} & $7\text{B}$ & 64 & $33.89$ & $49.20$ & $18.80$ & $58.20$ & $56.30$ & $43.27$\\
    Qwen-2.5-VL \cite{bai2025qwen25vl} & $7\text{B}$ & <=768 & $47.44$ & $59.20$ & $16.40$ & $56.00$ & $65.10$ & $48.82$\\ 
    InternVL-3 \cite{zhu2025internvl3} & $8\text{B}$ & $32$ & $49.33$ & $60.80$ & $26.50$ & $61.18$ & $59.08$ & $51.37$\\\midrule
    \textit{Reasoning Models} & & & & & & \\ \midrule
    Video-RTS \cite{wang2025videorts} & $7\text{B}$ & $128$ & $52.70$ & $66.40$ & $-$ & $63.00$ & $56.60$ & $-$\\
    Video-R1 \cite{feng2025video} & $7\text{B}$ & $128$ & $47.00$ & $64.00$ & $26.80$ & $64.30$ & $57.60$ & $51.94$ \\
    VideoChat-R1 \cite{li2025videochatr1} & $7\text{B}$ & $128$ & $52.00$ & $64.80$ & $26.50$ & $64.10$ & $54.30$ & $52.34$\\
    VideoChat-R1.5 \cite{yan2025videochatr1.5} & $7\text{B}$ & $128$ & $50.00$ & $67.00$ & $25.80$ & $64.80$ & $53.60$ & $52.24$\\
    Video-RFT \cite{wang2025videorft} & $7\text{B}$ & $128$ & $48.10$ & $66.70$ & $25.70$ & $64.10$ & $57.00$ & $52.32$\\
    Video-R2 \cite{maaz2025videor2} & $7\text{B}$ & $128$ & $50.80$ & $67.40$ & $28.40$ & $63.80$ & $59.20$ & $53.92$\\
    \midrule
    \textit{Test-Time Adaptation} & (Ours) & & & & & & \\ \midrule
    $TTRV^{*}$ (w Qwen2.5-VL) \cite{singh2025ttrv} & $7\text{B}$ & $32$ & $45.89$ & $64.48$ & $25.50$ & $59.26$ & $57.07$ & $50.46$ \\
    \rowcolor{lightblue}
    TTA-Vid (w Qwen2.5-VL) & $7\text{B}$ & $32$ & $49.44$ & $65.12$& $25.70$ & $61.14$ & $57.81$ & $51.84$\\
    \rowcolor{lightblue}
    TTA-Vid (w InternVL-3) & $8\text{B}$ & $32$ & $\textbf{53.66}$ & $65.60$ & $\textbf{29.80}$ & $\textbf{65.11}$ & $\textbf{61.48}$ & $\textbf{55.13}$\\

    \bottomrule
    \end{tabular}
    }
    \vspace{2mm}
    \caption{\textbf{Performance Comparison.} We compare TTA-Vid with the state-of-the-art methods on two instructional video question-answering tasks. The best results are highlighted in \textbf{bold}. For fair comparison, we repurpose TTRV with by training it in multi-frame setup on four frames.
    }
    \label{tab:maintable}
    \vspace{-10mm}
\end{table*}

\subsection{Benchmarks}

We evaluate the proposed method on five challenging and diverse video question-answering benchmarks: VideoMMMU \cite{hu2025videommmu}, MMVU \cite{zhao2025mmvu}, SciVideoBench \cite{deng2025scivideobench}, VideoMME \cite{fu2025videomme}, and LongVideoBench \cite{wu2024longvideobench}. 
Among these, VideoMMMU, MMVU and SciVideoBench are instructional videos, while  VideoMME focuses on generalized understanding and LongVideoBench focuses on long video reasoning problems. \textbf{VideoMMMU} \cite{hu2025videommmu} is a multimodal, multi-disciplinary benchmark that assesses LMM's ability to acquire and utilize knowledge from videos. 
The dataset has three splits: Perception, Comprehension, and Adaptation.
Each split contains 300 QA pairs, and a total of 900 QA pairs in the full benchmark. Perception questions assess the ability to perceive information from videos, Comprehension questions assess the ability to understand knowledge presented in videos, and Adaptation questions assess the ability to adapt video knowledge to new scenarios. \textbf{MMVU} \cite{zhao2025mmvu} is a comprehensive expert-level, multi-discipline benchmark which contains expert-annotated questions spanning 27 subjects across four core disciplines: Science, Healthcare, Humanities, Engineering, and Social Sciences. We test on the val split of MMVU on a multiple-choice QA format, which contains 625 QA pairs that require expert-level reasoning on complex videos. \textbf{SciVideoBench} \cite{deng2025scivideobench} is a benchmark with consists of 1000 MCQ questions from scientific experimental videos which span over 25 specialized academic subjects. \textbf{VideoMME} \cite{fu2025videomme} is a comprehensive evaluation benchmark consisting of videos from short to long videos (avg 17 min). We use the standard split of VideoMME which contains 2700 expert-labeled QA pairs designed for both perception and reasoning tasks. \textbf{LongVideoBench} \cite{wu2024longvideobench} is a videoqa benchmark that highlights referred reasoning questions, which are dependent on long frame inputs. We test on the validation split with 1337 reasoning questions.

\subsection{Implementation Details}
For the test-time adaptation, we consider a batch of randomly sampled 32 video question-answer pairs of a dataset. 
For training, we sample $K=4$ subsets from each video question-answer pair, consisting of $F=4$ frames, and do a step-by-step reasoning over $N=8$ rollouts per subset using a temperature of 1.0. %
We set the maximum prompt length to 7524 and the maximum response length to 1024 tokens.  
We set $\alpha=0.75$ in the final reward function (Equation~\ref{eq:reward_final}). 
We set the frame selection learning rate $\eta_{fs}=3$ (Equation~\ref{eq:mab}) for all our experiments.   
For hyperparameters, we use cosine learning rate schedule with a peak value of $5 \times 10^{-7}$ and adopt the AdamW optimizer for the policy model.
We train on this batch for five epochs. 
At test time, we test the model on the full test-set. 
The learned frame distribution for the samples in training set are averaged and  used as global frame distribution. 
For the final inference step, we sample 32 frames from each video 
and generate the answer based on the updated model.  
All experiments were conducted on 4 $\times$ NVIDIA A100 40GB GPUs.

\subsection{Comparison to State-of-the-Art}
In Table \ref{tab:maintable}, we compare TTA-Vid with proprietary MLLMs, open-source general-purpose multimodal models, and recent video reasoning LLMs across five benchmarks. When applied to InternVL3-8b and Qwen2.5VL-7b, TTA-Vid consistently improves the base backbones, achieving $55.13$ and $51.84$ average accuracy respectively. In particular, TTA-Vid (InternVL-38b) improves over its base model by $+3.76$ points (51.37 → 55.13), with gains across all the datasets. Compared to the open-source non-reasoning models such as LLaVA-OneVision, and InternVL, our method demonstrates clear advantages while maintaining the parameter scale. Although proprietary models like GPT-4o \cite{hurst2024gpt} remains stronger overall, TTA-Vid substantially narrows the gap. We further compare against video specific reasoning models, including Video-R1, VideoChat-R1, VideoChat-R1.5, Video-RFT, Video-R2 and VideoRTS. These approaches adopt multi-stage SFT+RL pipelines with substantial supervision: for example, VideoRFT was trained with over 100K CoT samples followed by more than 300K RL samples. Despite such extensive supervision, these models achieve average scores between $51.94$ and $53.92$, inferior compared to TTA-Vid ($55.13$). Our method requires only 32 unlabeled samples at test time, without any CoT/RL annotations, demonstrating the effectiveness of test-time adaptation and even surpass heavily trained video reasoning models. We repurpose TTRV which is trained with four frames, and is tested with 32 frames. TTRV improves the baseline scores on datasets like MMVU (avg. score from $48.82$ → $50.46$). On the other hand, TTA-Vid with batch-wide frequency reward-based RL, and MAB further boost the performance, achieving the best scores compared to the baseline and TTRV on all the datasets (avg. score from $48.82$ → $51.84$).

\subsection{Ablation Studies}
\begin{table*}[t]
    \centering
    \resizebox{1.0\textwidth}{!}{  %
        \begin{tabular}{@{}cccccccccc@{}}
        \toprule
        \multirow{2}{*}{TTRL} & \multirow{2}{*}{\shortstack{MAB\\Rewards}}  & \multirow{2}{*}{\shortstack{Gen. Frame\\Sampling}} & \multicolumn{4}{c}{VideoMMMU} & \multirow{2}{*}{MMVU} \\
        \cmidrule(lr){4-7}
        & &  & Perception & Comprehension & Adaptation & Avg & \\
        \midrule
        $\times$ & $\times$ &  $\times$ & 66.33 & 42.33 & 39.33 & 49.33 & 60.80 \\
        $\checkmark$ & $\checkmark$ &  $\times$ & 73.33 & 46.66 & 37.00 & 52.33 & 64.80\\
        \rowcolor{lightblue}$\checkmark$ & $\checkmark$ & $\checkmark$ & 74.33  & 48.66 & 38.00 & 53.66 & 65.60 \\
        \bottomrule
        \end{tabular}
    }
    \vspace{2mm}
    \caption{\textbf{Component ablation on InternVL3-8B.} Models are trained on subsets of each dataset for multiple epochs and evaluated on the full test sets, with results averaged across runs. Combining all components (highlighted) achieves the best overall performance across benchmarks. %
    }
    \label{tab:ablation}
    \vspace{-8mm}
\end{table*}

In this section, we present a comprehensive set of ablation studies to analyze the behavior and robustness of our method. We perform the following ablations: component-wise analysis with frame sampling (Section~\ref{ablation:componen_ablation}), full TTA vs. generalized setting (Section~\ref{ablation:gen_vs_full}), cross-dataset generalization (Section~\ref{ablation:cross_dataset_gen}), varying training sizes and sampling strategies (Sections~\ref{ablation:number_of_training_samples}–\ref{ablation:frame_sampling_type}), and ground-truth vs. majority-vote training (Section~\ref{ablation:maj_vs_gt}).

\definecolor{firstrow}{RGB}{230,245,255}   %
\definecolor{secondrow}{RGB}{210,235,255}  %

\begin{table*}[t]
\centering
\small
\setlength{\tabcolsep}{6pt}
\resizebox{\linewidth}{!}{
\begin{tabular}{lccccc}
\toprule
\multirow{2}{*}{Setup} 
& \multicolumn{4}{c}{VideoMMMU} 
& \multirow{2}{*}{MMVU} \\
\cmidrule(lr){2-5} & Perception & Comprehension & Adaptation & Avg &  \\
\midrule
InternVL3-2B & $47.33$ & $28.67$ & $28.33$ & $34.77$ & 50.72 \\
\rowcolor{firstrow} Full TTA & $48.23$ & $29.27$ & $31.14$ & $36.21$ & $53.45$ \\
\rowcolor{secondrow} Generalized TTA & $\textbf{48.33}$ & $\textbf{31.33}$ & $\textbf{32.00}$ & $\textbf{37.22}$ & $\textbf{54.40}$ \\
\bottomrule
\end{tabular}
}
\vspace{2mm}
\caption{\textbf{Full vs. generalized test-time adaptation.} TTA-Vid is trained on multiple subsets of 32 samples and evaluated on the same batch (Per-batch Evaluation) or on unseen batches (Generalized). Results are reported for InternVL3-2B.}
\label{tab:per_batch_training}
\vspace{-5mm}
\end{table*}

\definecolor{diagcolor}{RGB}{235,235,235}      %
\definecolor{bestcolor}{RGB}{220,235,255}      %

\begin{table}[t]
\centering
\resizebox{1.0\textwidth}{!}{%

\begin{tabular}{lcccc}
\toprule
\textbf{Train} $\downarrow$ / \textbf{Test} $\rightarrow$
& VideoMMMU
& MMVU 
& SciVideoBench 
& LongVideoBench \\
\midrule

Base InternVL3-8b 
& 49.33
& 60.80
& 26.50 
& 59.08 \\ \midrule

VideoMMMU
& \cellcolor{diagcolor}51.48 
& 62.08 
& \cellcolor{bestcolor}30.20 
& 60.81 \\

MMVU 
& \cellcolor{bestcolor}53.66 
& \cellcolor{bestcolor}65.60 
& 28.20 
& 60.95 \\

SciVid 
& 50.89 
& 61.28 
& \cellcolor{diagcolor}28.70 
& 59.98 \\

LVB 
& 50.55 
& 61.60 
& 28.80 
& \cellcolor{bestcolor}61.48 \\

\bottomrule
\end{tabular}
}
\vspace{2mm}
\caption{
Cross-dataset generalization of InternVL3-8B (32 frames, global averaging).
Rows denote training dataset and columns denote evaluation dataset.
\colorbox{bestcolor}{Best} indicates highest performance per test set.
\colorbox{diagcolor}{Diagonal} indicates in-domain results. 
}
\label{tab:cross_data_gen}
\vspace{-8mm}
\end{table}

\definecolor{lightyellow}{RGB}{255,250,220} 
\begin{table}[t]

\centering
\begin{minipage}[t]{0.45\textwidth}
\centering
\begin{tabular}{c|cc}
\toprule
\multirow{2}{*}{\#Samples} & \multicolumn{2}{c}{Accuracy (\%)} \\
\cmidrule(lr){2-3}
 & Perception & MMVU \\
\midrule
Qwen2.5-VL & 58.33 & 59.20 \\
$V=1$ & 67.00 & 61.28 \\
$V=4$ & 67.33 & 61.60 \\
$V=8$ & 67.67 & 61.28 \\
$V=16$ & 67.67 & 62.60 \\
\midrule
\rowcolor{lightblue}$V=32$ & 69.00 & 65.12 \\
\bottomrule
\end{tabular}
\vspace{2mm}
\caption{Effect of training sample size on model performance. Evaluation is conducted on the VideoMMMU-Perception split and the MMVU benchmark with TTA-Vid+Qwen2.5VL-7b. 
}
\label{tab:ts_ablation}
\vspace{-8mm}
\end{minipage}
\hfill
\begin{minipage}[t]{0.45\textwidth}
\centering
\begin{tabular}{lcc}
\toprule
\multirow{2}{*}{Type} & \multicolumn{2}{c}{Accuracy (\%)} \\
\cmidrule(lr){2-3}
 & Perception & MMVU \\
\midrule
InternVL3-8b & 66.33 & 60.80 \\ 
Random & 73.33 & 64.80 \\
CLIP & 73.00 & 65.28 \\ \midrule
\rowcolor{lightblue}(Ours) & 74.33 & 65.60 \\
\bottomrule
\end{tabular}
\vspace{2mm}
\caption{Effect of different generalized frame sampling strategies on VideoMMMU-Perception and MMVU with TTA-Vid+InternVL3-8b. The learned distribution performs slightly better than other baselines.}
\label{tab:frame_sampling}
\vspace{-8mm}
\end{minipage}
\end{table}

\paragraph{Component Ablation.}
\label{ablation:componen_ablation}
We first evaluate the contribution of both major components: the test-time adaptation via batched rewards and frame sampling. Table~\ref{tab:ablation} shows the performance as we ablate both components, comparing the baseline performance, as well as TTRL with batched rewards, and the full method combining both batched rewards with adaptive frame sampling. It shows that each component yields distinct gains across different benchmarks. Namely, TTRL with batched rewards provides an average improvement of $+3\%$ on the VideoMMU and $4\%$ on the MMVU dataset. Furthermore, adding frame sampling boosts the performance by $+4.33\%$ on VideoMMMU and $+4.8\%$ on MMVU.

\paragraph{Full vs Generalized TTA.}
\label{ablation:gen_vs_full}
We evaluate TTA-Vid under two settings. In the first setting (Full TTA), both training and testing are performed on the same batch of samples. In the second setting (Generalized TTA), the model is trained on a single batch and then evaluated on the entire test set. For the full method, we use the average frame distribution learned across videos during training. We use InternVL3-2B for this ablation with respect to computational runtime. The results are reported in Table~\ref{tab:per_batch_training}. We observe that per-batch training and testing already lead to consistent improvements over the baseline, indicating that the RL-based reward signal, together with our frame-sampling strategy, provides meaningful optimization even within a single batch. But interestingly, we achieve the strongest overall performance for the generalized case. This suggests that the adaptation process does not overfit to the specific batch used for training, but instead improves the model in a way that generalizes effectively to unseen data.

\paragraph{Cross-dataset generalization.} 
\label{ablation:cross_dataset_gen} 
In Table~\ref{tab:cross_data_gen}, we present the cross-dataset generalization results for InternVL3-8B. In this setup, each model is trained on 32 samples from a single dataset and then evaluated on all datasets, including its own. The diagonal entries denote in-domain performance, while off-diagonal entries reflect cross-dataset transfer. 
Overall, a model trained on any individual dataset achieves competitive and in most cases stable performance across other datasets, indicating strong generalization. Notably, several off-diagonal results surpass the corresponding in-domain (diagonal) scores (example: Perception (72.67 → 74.67) with Comprehension and Adaptation model, SciVideoBench (28.70 → 30.20) with Perception model), showing that the learned model is not limited to dataset-specific patterns. %

\paragraph{Different number of samples in the training set.} 
\label{ablation:number_of_training_samples}
In this experiment, Table~\ref{tab:ts_ablation}, we analyze the effect of varying the number of training samples $V=\{1, 4, 8, 16, 32\}$ on model performance. Specifically, we train QwenVL-7B using different numbers of samples per training setup. We show results on the VideoMMMU-Perception split and the MMVU dataset to assess consistency across benchmarks. It can be seen that even with a batch size of one, $V={1}$, the base model's accuracy increases from $58.33\%$ to $67.00\%$. Adding more samples to the training set furthermore, increases accuracy, with $V={32}$ obtaining $69.00\%$ accuracy. A similar trend is also observed for MMVU dataset.

\paragraph{Generalized frame sampling.} 
\label{ablation:frame_sampling_type}
In Table~\ref{tab:frame_sampling}, we show the results where all the experiments are conducted on the same model, i.e. trained with adaptive frame selection but with different frame sampling strategies during testing which are random sampling, CLIP score-based sampling, and our learned generalized frame sampling. We ablate with the two datasets: VideoMMMU-Perception split and MMVU dataset. It can be seen that CLIP based sampling helps MMVU ($65.28\%$) when compared to random sampling. However, the best scores are obtained when we use the learned frame distribution $74.33\%$ and $65.60\%$ on VideoMMMU-Perception and MMVU datasets, respectively.

\paragraph{Majority answer vs Ground Truth answer vs Self-consistency.}
\label{ablation:maj_vs_gt}
In this experiment on the MMVU dataset, we replace the majority-vote answer used for reward computation with the ground-truth answer during training to assess the robustness of the majority-based reward formulation. We train the Qwen2.5VL-7B model under both settings. The model trained using ground-truth answers achieves an accuracy of 60.96\%, while the model trained with majority-vote answers attains 65.11\% accuracy. These results suggest that majority-vote rewards lead to better generalization, rather than overfitting to ground-truth answers, and contribute to improved test-time adaptability. To compare the proposed method against a common test-time scaling method, we test self-consistency on MMVU dataset. It achieves a score of $62.72\%$, whereas our proposed method reaches $65.60\%$, outperforming self-consistency. Moreover, self-consistency requires an increasing number of forward passes as the dataset size grows, while the proposed method needs to be trained only once for a batch.

\begin{figure*}[t]
    \centering
    \includegraphics[width=0.99\textwidth]{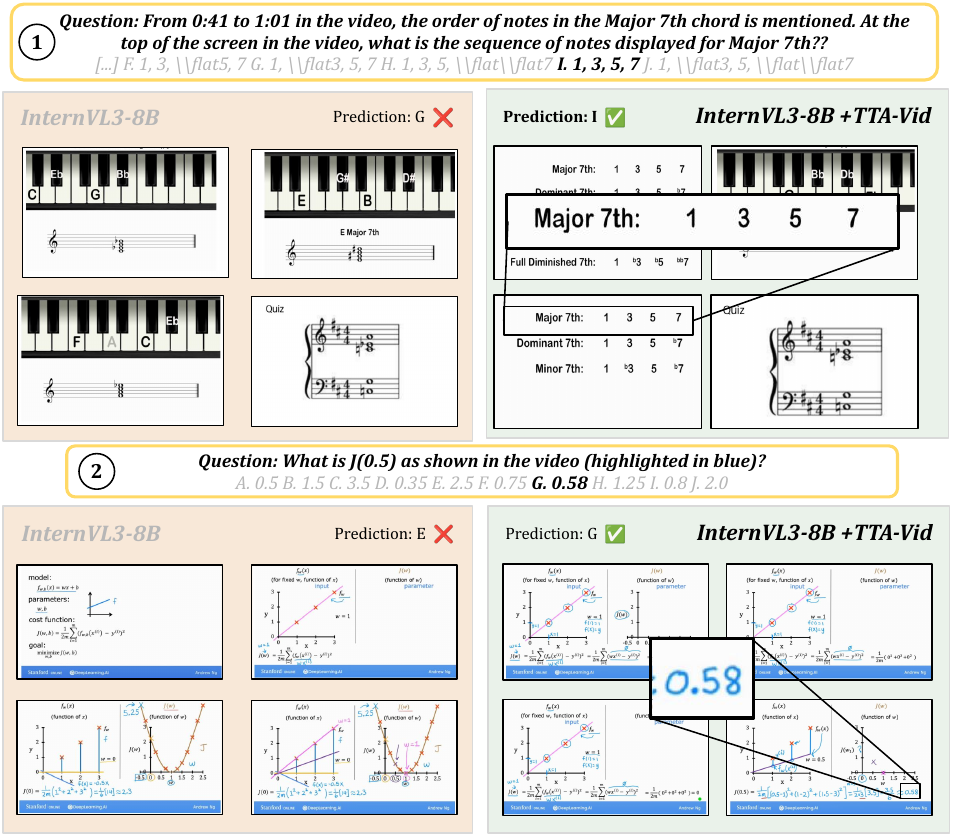}
    \caption{\textbf{Qualitative comparison of frame selection strategies.} Random sampling (left) versus our learned selection (right) on two VideoMMMU (Perception) examples. \textbf{(1)}: Musical question which requires the model to identify the sequence of notes displayed at a certain timestamp of the video. Our method selects the critical frame (correct answer I), while random sampling misses it (predicts G). \textbf{(2)}: Accounting question requiring localization of the value. Our method identifies the highlighted value in blue to be ({0.58}, option G), while random sampling fails (predicts {2.5}, option E).}
    \label{fig:qualitative}
    \vspace{-6mm}
\end{figure*}

\subsection{Qualitative Analysis}
To illustrate the effectiveness of our learned frame selection, we compare the combination of the test-time adaptation and frames selection by our optimization-based approach against random sampling. Figure~\ref{fig:qualitative} presents two representative examples from VideoMMMU-Perception. The frames shown in the light orange background are from when frames are sampled randomly, and the frames shown in the green background are when frames are sampled with the learned distribution. In the first example, the question asks to identify the correct sequence of notes displayed in a  specific timestamp, requiring both temporal localization (identifying the frames in ``From $0$:$41$ to $1$:$01$ in the video'') and some contextual understanding that chord corresponds to a sequence of notes. Our adapted model retrieves the relevant frame showing the correct note order "$1$, $3$, $5$, $7$ (option "I."), whereas the baseline model fails in capturing the correct frame (predicting "G."). In the second example,  the question requires identifying the specific textual content visible on the video: ``What is $J(0.5)$ as shown in the video (highlighted in blue)?''. The adapted model successfully selects the frames that lead to the correct answer, i.e. first identifying the frames in which $J(0.5)$ is highlighted in blue color (as specified in the question), followed by continuous selection of important frames, to finally select the frame that has the answer in it, which is option G same as the ground truth. In contrast, the random sampling misses these key frames, leading to an incorrect baseline prediction (J). These examples demonstrate that the proposed adaptive frame selection, coupled with the test-time adaptation with reinforcement learning, learns to prioritize frames containing task-relevant information.

\section{Conclusion}
In this work, we introduced TTA-Vid, the first test-time reinforcement learning framework for long instructional videos, where rewards are extracted on-the-fly from unlabeled test data. Our approach addresses the challenge of adapting vision-language models to notoriously difficult instructional video content at inference time by leveraging consistency-based self-reinforcement signals: we compute a frequency-based reward from agreement among generated answers across diverse frame subsets, and optimize a multi-armed bandit formulation using these test-time reinforcement learning signals to learn interpretable frame importance distributions. 
Our extensive evaluation across five different datasets demonstrates consistent improvements over strong baseline models, with gains up to $4.33\%$ on VideoMMMU. Beyond empirical gains, TTA-Vid enhances video understanding abilities without annotations, pointing to test-time optimization through reinforcement learning as a powerful paradigm for bridging pre-trained vision-language models and downstream instructional video understanding.

\section*{Acknowledgements}
Soumya Jahagirdar is funded by the European Research Council (ERC) under the Starting Grant \emph{GraViLa} (101117556). Edson Araujo is funded by European Union's Horizon Europe research and innovation programme under grant agreement number 101214398 (ELLIOT). This work is supported by the T\"ubingen AI Center. This research is in part supported by the MIT-IBM Watson AI Lab.

\bibliographystyle{splncs04}
\bibliography{main}

\appendix
\clearpage
\setcounter{page}{1}

\section*{Supplementary Material}
In this appendix, we provide additional experimental results, detailed experimental settings, and qualitative examples. In Section~\ref{subsec:knablation}, we report performance across varying subset sizes and rollouts. Section~\ref{subsec:training_epoch} presents the performance progression per training epoch on the MMVU dataset. 
In Section~\ref{subsec:frame_selection}, we evaluate multiple frame sampling strategies. Section~\ref{subsec:uniforminit_vs_clipinit} examines different weight initialization methods for frame weight updates. In Section~\ref{subsec:longvbench_category_performance}, we report category-wise performance on LongVideoBench and SciVideoBench dataset. In Section~\ref{subsec:continual_training}, we experiment with continual training across multiple datasets. In Section~\ref{subsec:motivation}, we explain why majority voting works and how it enables frame selection with MAB. In Section~\ref{subsec:runtime_analysis}, we show computational cost, and finally, in Section~\ref{subsec:quals}, we show a few qualitative examples along with some failure cases.

\section{Varying number of Subsets (K) and Rollouts (N)}
\label{subsec:knablation}
To better understand the sensitivity of our method to its sampling configuration, we analyze the effect of varying the number of subsets K and rollouts N. In this experiment, we vary the number of Subsets from $K=\{4,8\}$ and $N=\{8,16,32\}$. We experiment this on MMVU dataset with Qwen2.5VL-7B as the base model. From Table~\ref{tab:knablation}, it can be seen that, for $K=4$ subsets, the best results are obtained with $N=32$ rollouts, with $0.48\%$ increase in performance compared to the default setting ($K=4$, $N=8$). For $K=8$ subsets, the best performance is observed with $N=16$ rollouts. This experiment suggests the robustness of the proposed method with different Subsets ($K$) and Rollouts ($N$).

\section{Impact of number of training epochs}
\label{subsec:training_epoch}

To examine how training duration affects performance, we analyze the impact of the number of training epochs by training Qwen2.5VL-7B on the MMVU \cite{zhao2025mmvu} dataset for five epochs. We save the model checkpoint after each epoch and evaluate the performance of the model. Table~\ref{tab:training_epoch} reports the accuracy obtained at different epochs.
Starting from a baseline model, the model shows a substantial improvement after the first epoch, reaching 64.48\%. The performance then continues to improve gradually with additional training, increasing to 64.64\% at Epoch 2 and 64.80\% at Epoch 3. The best performance of 65.12\% is achieved at Epoch 4. Overall, these results demonstrate a consistent upward trend in accuracy across epochs, suggesting that additional training epochs allow the model to better adapt to the dataset while providing stable performance gains.

\begin{table*}[t]
\centering
\begin{minipage}{0.40\linewidth}
\centering
\resizebox{0.95\linewidth}{!}{
\begin{tabular}{@{}ccc@{}}
\toprule
Subsets (K) & Rollouts (N) & MMVU \\
\midrule
\rowcolor{lightblue}
4 & 8 & 65.12 \\
4 & 16 & 65.28 \\
4 & 32 & 65.60 \\ 
\midrule
8 & 8 & 64.96 \\
8 & 16 & 65.60 \\
8 & 32 & 64.96 \\  
\bottomrule
\end{tabular}
}
\vspace{2mm}
\caption{\textbf{Component ablation in TTA-Vid.} Performance with different subsets (K) and rollouts (N) using Qwen2.5VL-7B. The setup used in the paper is \colorbox{lightblue}{\textit{highlighted}}.}
\label{tab:knablation}
\end{minipage}
\hfill
\begin{minipage}{0.29\linewidth}
\centering
\resizebox{0.9\linewidth}{!}{
\begin{tabular}{@{}lc@{}}
\toprule
Epoch & MMVU\\
\midrule
Qwen2.5VL-7B & 59.20 \\
\midrule
Epoch 0 & 64.48 \\
Epoch 1 & 64.48 \\ 
Epoch 2 & 64.64 \\
Epoch 3 & 64.80 \\ 
\rowcolor{lightblue}
Epoch 4 & 65.12 \\  
\bottomrule
\end{tabular}
}
\vspace{2mm}
\caption{\textbf{Impact of Number of Training Epochs.} Performance across epochs on MMVU with Qwen2.5VL-7B. The setup used in the paper is \colorbox{lightblue}{\textit{highlighted}}.}
\label{tab:training_epoch}
\end{minipage}
\hfill
\begin{minipage}{0.24\linewidth}
\centering
\resizebox{0.9\linewidth}{!}{
\begin{tabular}{@{}lc@{}}
\toprule
Method & MMVU \\
\midrule
Random & 50.72 \\
Uniform & 53.92 \\
BLIP & 50.88 \\ 
AKS & 53.92 \\
\rowcolor{lightblue}
TTA-Vid & 55.04 \\ 
\bottomrule
\end{tabular}
}
\vspace{2mm}
\caption{\textbf{Frame selection baselines}. MMVU dataset with InternVL3-2b as the base model.}
\label{tab:frame_selection_baseline}
\end{minipage}
\end{table*}

\section{Frame selection}
\label{subsec:frame_selection}

Since TTA-Vid relies on selecting informative frames under a limited frame budget, it is important to understand how different frame selection signals contribute to performance. We analyze this along two axes: the combination of CLIP-based scores and learned distributions (Table~\ref{tab:frame_selection}), and the impact of different temporal sampling strategies (Table~\ref{tab:1fps}).

\paragraph{CLIP and learned distribution weighting.}
In Table~\ref{tab:frame_selection}, we analyze different frame selection strategies at test time using the TTA-Vid + InternVL3-8B model. Specifically, we vary the contribution of CLIP-based frame scores ($W_{clip}$) and the learned frame distribution ($W_{dist}$).
For a small number of frames (four frames), the combination $0.2W_{clip} + 0.8W_{dist}$ achieves the best performance across all datasets, obtaining 53.55 on LongVideoBench, 57.48 on VideoMME, and 27.90 on SciVideoBench. Compared to using only CLIP scores ($W_{clip}=1.0$) or only the learned distribution ($W_{dist}=1.0$), this combination consistently improves performance. This suggests that combining CLIP-based relevance scores with the learned frame distribution provides a strong prior for selecting informative frames when the frame budget is limited. For a larger frame budget (32 frames), the optimal weighting varies across datasets. The model achieves the best performance on LongVideoBench when relying entirely on the learned distribution ($W_{dist}=1.0$), while VideoMME performs best when using only CLIP scores ($W_{clip}=1.0$). SciVideoBench achieves the highest accuracy with a mixed weighting ($0.8W_{clip} + 0.2W_{dist}$). These results indicate that while the learned distribution provides a strong general prior, CLIP-based scoring can still offer complementary information depending on the dataset characteristics.

\paragraph{Uniform sampling vs. 1 FPS.}
In Table~\ref{tab:1fps}, we further evaluate the robustness of TTA-Vid under a different sampling strategy by selecting frames at one frame per second (32 frames) instead of the default uniform sampling. Since the learned distribution is trained on 40 uniformly sampled frames, we interpolate the distribution to match the 1fps sampling. Although the performance slightly decreases across datasets (e.g., 65.60 $\rightarrow$ 64.96 on MMVU and 61.48 $\rightarrow$ 61.11 on LongVideoBench), the results remain competitive, demonstrating that TTA-Vid generalizes well to alternative frame sampling strategies.

\begin{table*}[t]
    \centering
    \resizebox{0.8\linewidth}{!}{
    \begin{tabular}{@{}c|cc|ccc@{}}
    \toprule
     \# frames & $W_{clip}$ & $W_{dist}$ & LongVideoBench & VideoMME & SciVideoBench\\
    \midrule
    4 & 1.0 & 0.0 & 52.35 & 56.48 & 27.70 \\
    4 & 0.8 & 0.2 & 52.58 & 56.88 & 27.50 \\
    4 & 0.5 & 0.5 & 52.80 & 57.11 & 27.80 \\
    4 & 0.2 & 0.8 & \textbf{53.55} & \textbf{57.48} & \textbf{27.90} \\
    4 & 0.0 & 1.0 & 52.05 & 55.51 & 27.80 \\ \midrule

    32 & 1.0 & 0.0 & 60.58 & \textbf{65.74} & 29.80 \\
    32 & 0.8 & 0.2 & 59.83 & 65.33 & \textbf{30.30} \\
    32 & 0.5 & 0.5 & 60.58 & 65.70 & 29.30 \\
    32 & 0.2 & 0.8 & 61.25 & 64.96 & 29.30 \\
    \rowcolor{lightblue}
    32 & 0.0 & 1.0 & \textbf{61.48} & 65.11 & 29.80 \\

    \bottomrule
    \end{tabular}
}  
    \vspace{2mm}
    \caption{\textbf{Frame sampling using CLIP and learned distribution.} In this Table, we analyze effect of using frame sampling based on the combination of CLIP score and learned distribution on InternVL3-8b as the base model. The setup used in the paper is \colorbox{lightblue}{\textit{highlighted}}.}\label{tab:frame_selection}
\end{table*}

\begin{table*}[t]
    \centering
    \resizebox{0.8\linewidth}{!}{
    \begin{tabular}{@{}l|cccc@{}}
    \toprule
     Frame Sampling & MMVU & LongVideoBench & VideoMME & SciVideoBench\\
    \midrule
    \rowcolor{lightblue}
    Uniform & 65.60 & 61.48 & 65.11 & 29.80 \\
    1 FPS & 64.96 & 61.11 & 64.00 & 29.40 \\
    \bottomrule
    \end{tabular}
}  
    \vspace{2mm}
    \caption{\textbf{Uniform sampling vs. 1 FPS.} Comparison of TTA-Vid performance using two frame sampling strategies: (a) selecting 32 frames from 40 uniformly sampled frames and (b) sampling one frame per second from the video (at least and up to 32 frames). The setup used in the paper is \colorbox{lightblue}{\textit{highlighted}}.}\label{tab:1fps}
\end{table*}

\definecolor{evalcell}{RGB}{220,235,255}
\definecolor{darkgreen}{RGB}{0,140,60}

\begin{table}[t]
\centering
\resizebox{0.9\linewidth}{!}{
\begin{tabular}{c@{\hspace{8pt}}cccc@{\hspace{10pt}}c}
\toprule
\multirow{2}{*}{Stage}
& \multicolumn{4}{c}{Training Dataset}
& \multicolumn{1}{c}{Test Acc. \tiny{(in\%)}} \\
\cmidrule(r{8pt}){2-5}
& MMVU & VideoMMMU & SciVid & Video-MME  & TTA-Vid \\
\midrule
1 & \cellcolor{evalcell}\checkmark & - & - & -  & 64.48 \textcolor{darkgreen}{\tiny($+5.28\%$)} \\
2 & \checkmark & \cellcolor{evalcell}\checkmark & - & - & 48.83 \textcolor{darkgreen}{\tiny($+1.39\%$)} \\
3 & \checkmark & \checkmark & \cellcolor{evalcell}\checkmark & - &  25.80 \textcolor{darkgreen}{\tiny($+9.40\%$)} \\
4 & \checkmark & \checkmark & \checkmark & \cellcolor{evalcell}\checkmark  & 61.22 \textcolor{darkgreen}{\tiny($+5.22\%$)} \\
\bottomrule
\end{tabular}
}
\vspace{2mm}
\caption{\textbf{Continual training across datasets.} Each stage initializes from the previous model and is further trained with an additional dataset. \textit{We train the model on each stage only for one epoch.} Colored cells indicate the dataset used for evaluation at each stage.}
\label{tab:continual_exp}
\end{table}

\section{Uniform initialization vs. CLIP score initialization}
\label{subsec:uniforminit_vs_clipinit}

As explained in Section~\ref{sec:tta}, we initialize the weights of the multi-armed bandit using CLIP scores computed for each question–frame pair. To further evaluate the robustness of the proposed method, we consider the scenario where CLIP scores are unavailable and instead initialize the distribution uniformly. In this case, frames are initially sampled randomly rather than according to the CLIP-based prior. With uniform initialization, the Qwen2.5VL-7B model achieves an accuracy of $65.44\%$ on the MMVU dataset, compared to $65.60\%$ when initialized with CLIP scores. The small performance gap indicates that the proposed batched-reward optimization can learn an effective frame distribution regardless of the initialization strategy. These results demonstrate that while CLIP scores provide a useful prior, the method remains robust even when such information is not available. 

\begin{figure*}[t]
    \centering

    \begin{subfigure}[t]{\linewidth}
        \centering
        \includegraphics[width=\linewidth]{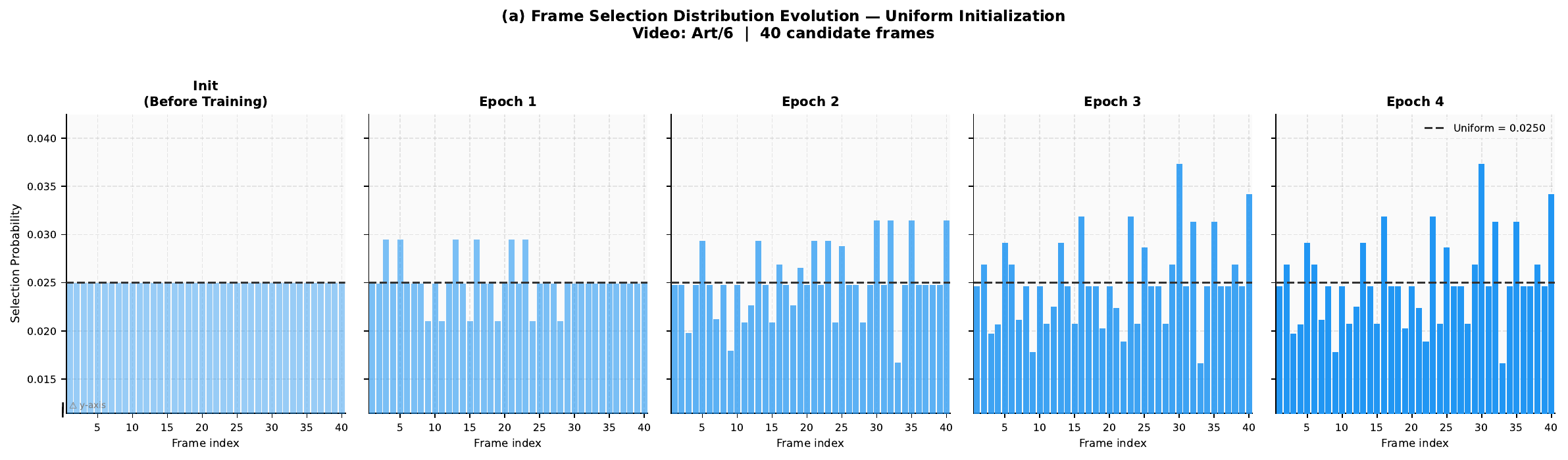}
        \caption{\textbf{Uniform initialization distribution progression over epochs.}}
        \label{fig:plot_frame_dist_example_unif}
    \end{subfigure}
    \begin{subfigure}[t]{\linewidth}
        \centering
        \includegraphics[width=\linewidth]{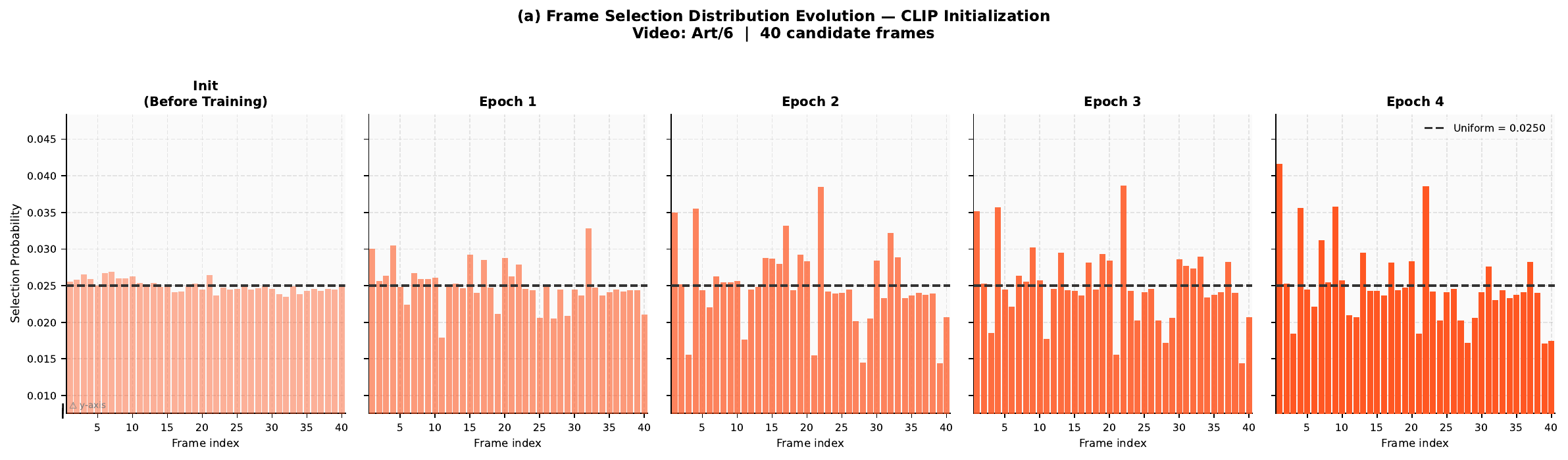}
        \caption{\textbf{CLIP initialization distribution progression over epochs.}}
        \label{fig:plot_frame_dist_example_clip}
    \end{subfigure}

    \caption{\textbf{Distribution evolution.} Distribution progression for uniformly sampled 40 frames, over epochs for a single example with two types of initialization: uniform initialization vs CLIP initialization.}
    \label{fig:frame_distributions}
\end{figure*}

To further illustrate this behavior, we visualize the evolution of the frame selection distribution over training epochs in Figure~\ref{fig:frame_distributions}. The figure shows the distribution progression for a representative example with 40 uniformly sampled frames under both uniform initialization and CLIP-based initialization. When initialized uniformly (Figure~\ref{fig:plot_frame_dist_example_unif}), all frames begin with equal probability. As training progresses, the batched-reward optimization gradually shifts probability towards a subset of informative frames, resulting in a more peaked and structured distribution by later epochs. In contrast, when initialized with CLIP scores (Figure~\ref{fig:plot_frame_dist_example_clip}), the initial distribution already reflects frame relevance estimated by CLIP. Training further refines this distribution, amplifying important frames while suppressing less informative ones. Notably, despite the different starting points, both initialization strategies converge to informative distributions over time. This analysis supports the quantitative results in Table~\ref{tab:training_epoch}, demonstrating that the proposed optimization procedure can reliably discover informative frame subsets regardless of the initialization strategy. 

\section{Category performance for LongVideoBench and SciVideoBench}
\label{subsec:longvbench_category_performance}
We provide a category-wise breakdown of the results on LongVideoBench \cite{wu2024longvideobench} and SciVideoBench \cite{deng2025scivideobench}. These two benchmarks offer rich category annotations that allow for a fine-grained analysis beyond aggregate scores: LongVideoBench covers diverse topic categories and video durations, while SciVideoBench spans multiple scientific disciplines and question types. Analyzing performance at this granularity helps identify the specific settings where TTA-Vid provides the largest improvements over the base model.

\begin{figure*}[t]
    \centering
    \includegraphics[width=\linewidth]{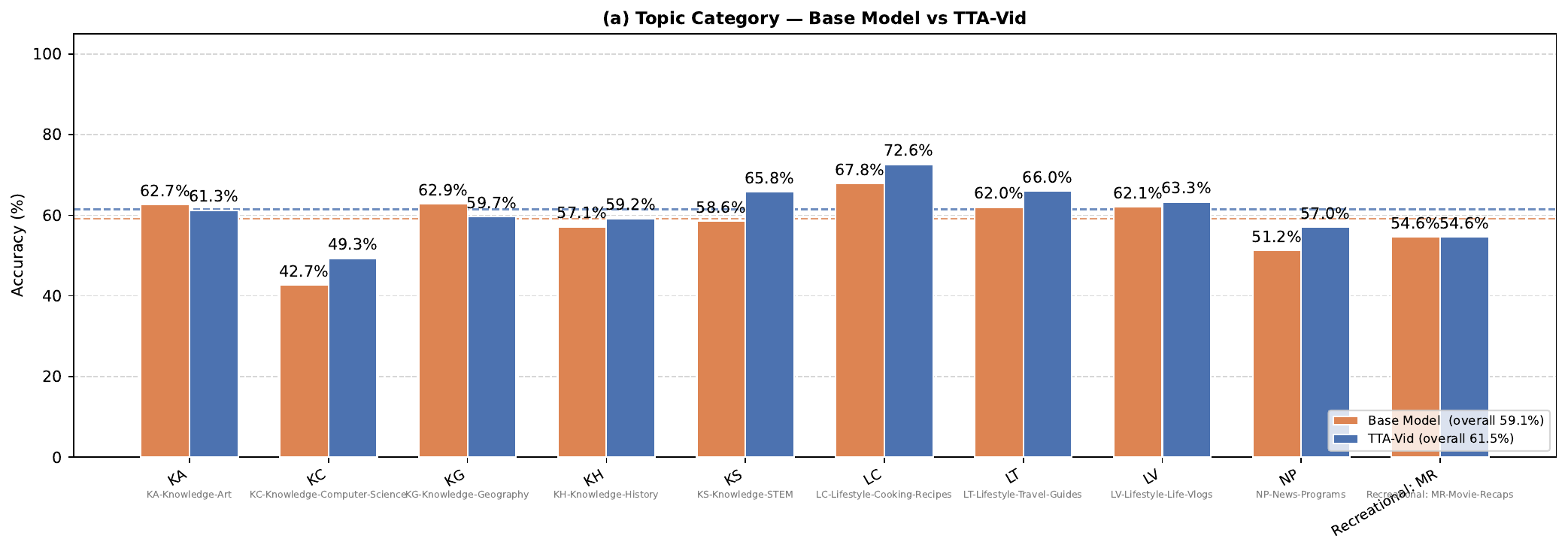}
    \caption{\textbf{Performance per category: LongVideoBench} Orange and blue dotted lines correspond to the overall performance of base model and TTA-Vid respectively. LongVideoBench dataset has multiple categories. The highest gain over the base model are observed for the following categories: Computer Science, STEM, New programs.}
    \label{fig:lvb_plot_a_topic_category_base_vs_ft}
\end{figure*}

\begin{figure*}[t]
    \centering
    \includegraphics[width=\linewidth]{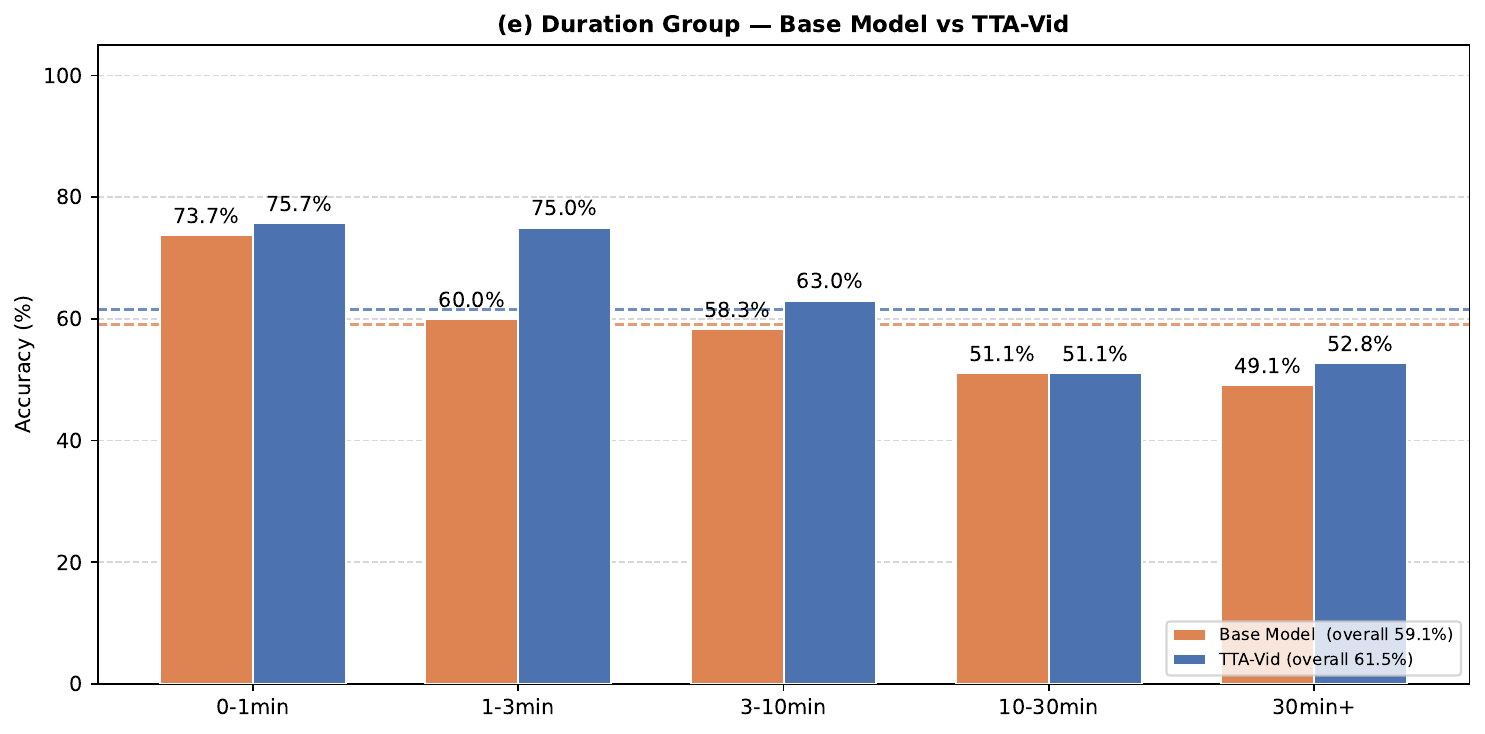}
    \caption{\textbf{Performance based on duration of the video: LongVideoBench} Orange and blue dotted lines correspond to the overall performance of base model and TTA-Vid respectively. The highest gains are observed for videos with duration of 1-3 minutes. TTA-Vid also works well with very long videos, i.e. over 30 minutes.}
    \label{fig:lvb_plot_e_duration_group}
\end{figure*}

\paragraph{LongVideoBench.}
LongVideoBench contains multiple topic categories that allow for a fine-grained analysis of model performance. As shown in Figure~\ref{fig:lvb_plot_a_topic_category_base_vs_ft}, TTA-Vid outperforms the base model in 7 out of 10 categories, indicating that the proposed method consistently improves performance across diverse video topics. We further analyze the performance with respect to video duration in Figure~\ref{fig:lvb_plot_e_duration_group}, where we compare the base and TTA-Vid across different video length groups. The results show that the TTA-Vid achieves improvements across all duration ranges. This demonstrates that the proposed approach generalizes well to videos of varying lengths and is effective even for long-duration videos, highlighting the robustness of the proposed method.

\begin{figure*}[t]
    \centering

    \begin{subfigure}[t]{\linewidth}
        \centering
        \includegraphics[width=\linewidth]{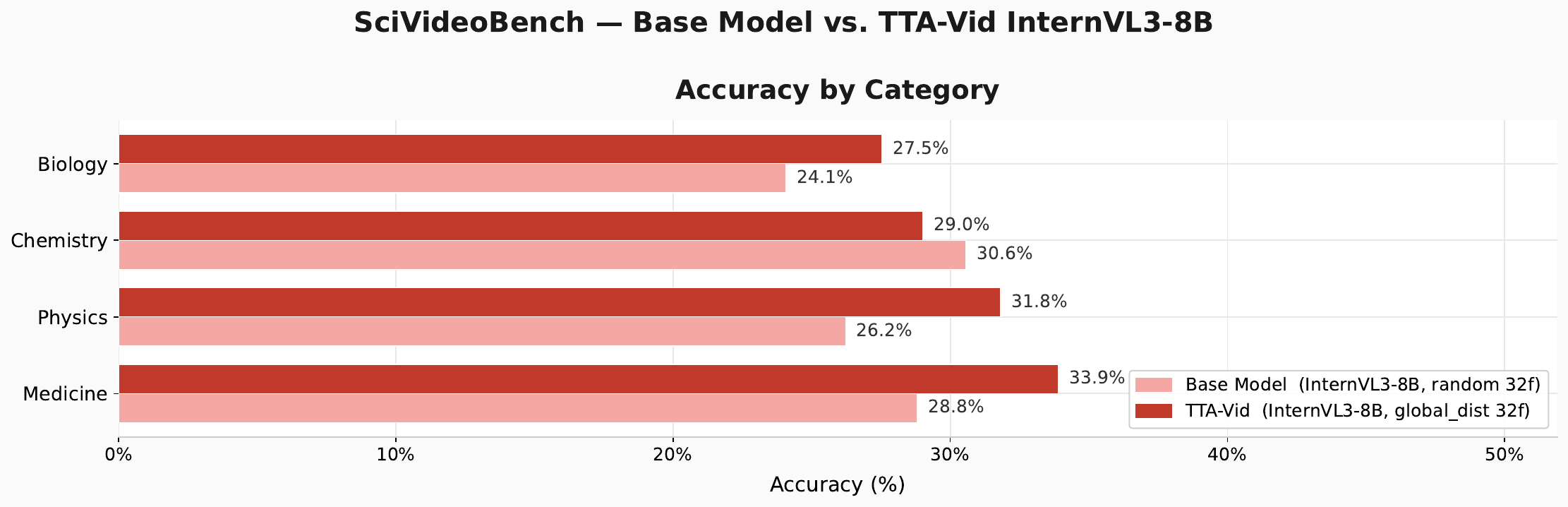}
        \caption{\textbf{Category.}}
        \label{fig:scivideobench_base_vs_finetuned_category}
    \end{subfigure}
    \begin{subfigure}[t]{\linewidth}
        \centering
        \includegraphics[width=\linewidth]{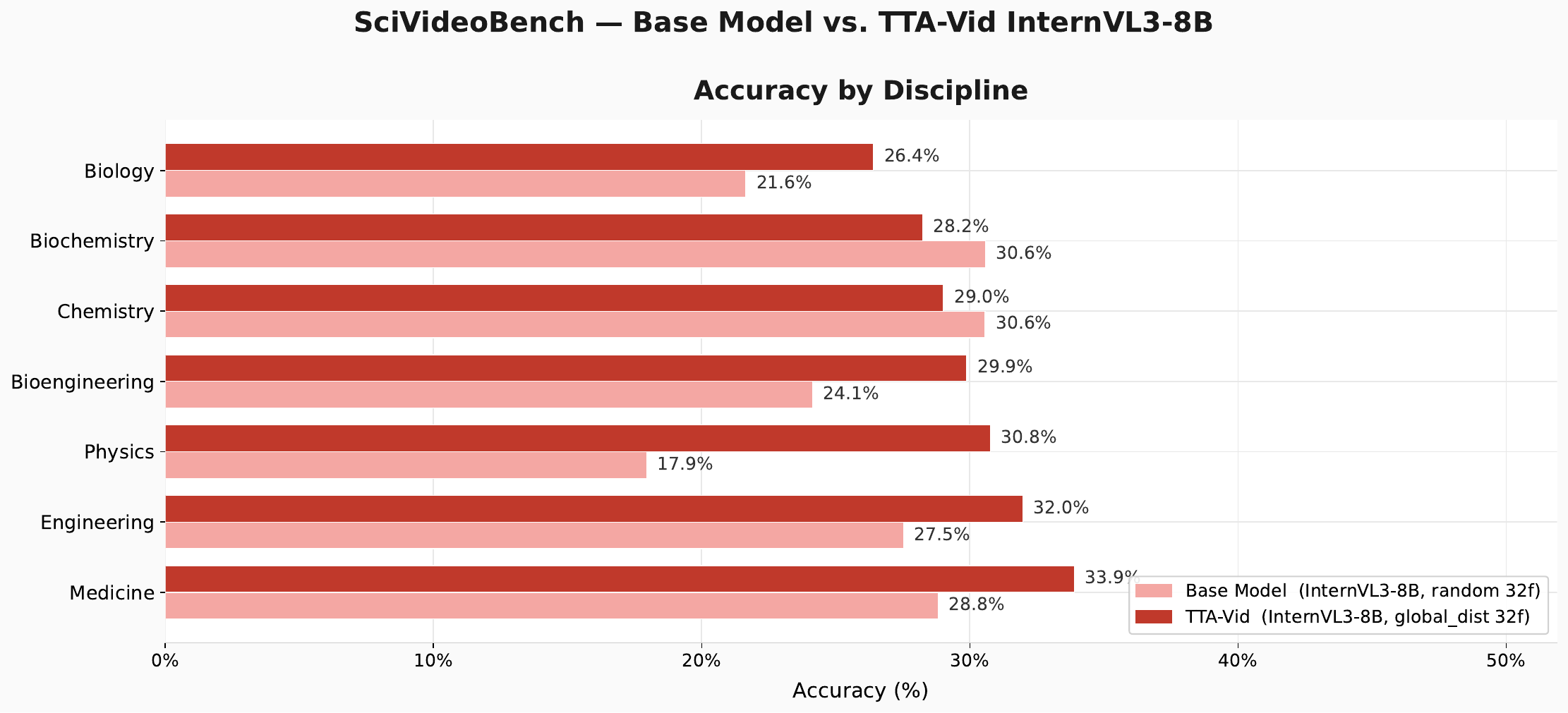}
        \caption{\textbf{Disciplines.}}
        \label{fig:scivideobench_base_vs_finetuned_discipline}
    \end{subfigure}
    \begin{subfigure}[t]{\linewidth}
        \centering
        \includegraphics[width=\linewidth]{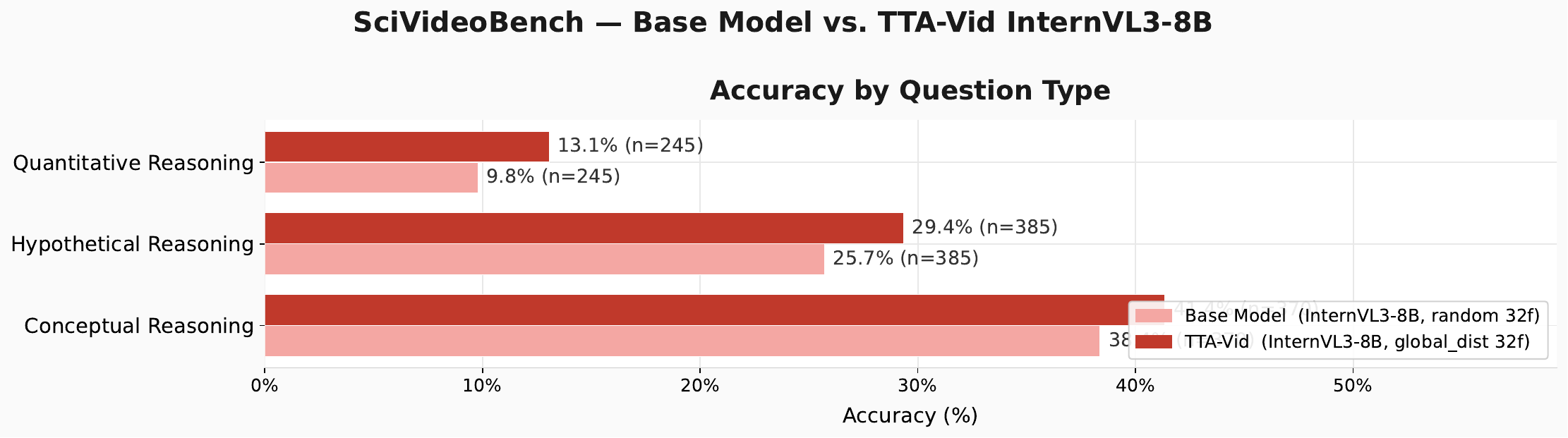}
        \caption{\textbf{Question type.}}
        \label{fig:scivideobench_base_vs_finetuned_question_type}
    \end{subfigure}
    \caption{\textbf{SciVideoBench performance analysis.} Comparison between the base InternVL3-8B model and TTA-Vid model with the learned frame selection distribution. We report accuracy across (a) high-level scientific categories, (b) specific scientific disciplines, and (c) question types. 
    }
    \label{fig:scivideobench}
\end{figure*}

\paragraph{SciVideoBench.}
We further analyze the performance of TTA-Vid+InternVL3-8b and base InternVL3-8b model on SciVideoBench dataset across different categories, disciplines, and question types in Figure~\ref{fig:scivideobench}. Across high-level categories (Figure~\ref{fig:scivideobench_base_vs_finetuned_category}), TTA-Vid consistently improves performance over the base model, with notable gains in Medicine (33.9\% vs.\ 28.8\%) and Physics (31.8\% vs.\ 26.2\%). Similar trends are observed when analyzing performance across scientific disciplines (Figure~\ref{fig:scivideobench_base_vs_finetuned_discipline}). The largest improvements appear in Physics and Bioengineering. We analyze results across question types in Figure~\ref{fig:scivideobench_base_vs_finetuned_question_type}. TTA-Vid shows improvements across all reasoning types, including quantitative reasoning (13.1\% vs.\ 9.8\%) and hypothetical reasoning (29.4\% vs.\ 25.7\%). These results suggest that the proposed approach improves the model’s ability to identify informative frames that support diverse forms of scientific reasoning in video-based question answering.

\begin{figure*}[t]
    \centering

    \begin{subfigure}[t]{\linewidth}
        \centering
        \includegraphics[width=\linewidth]{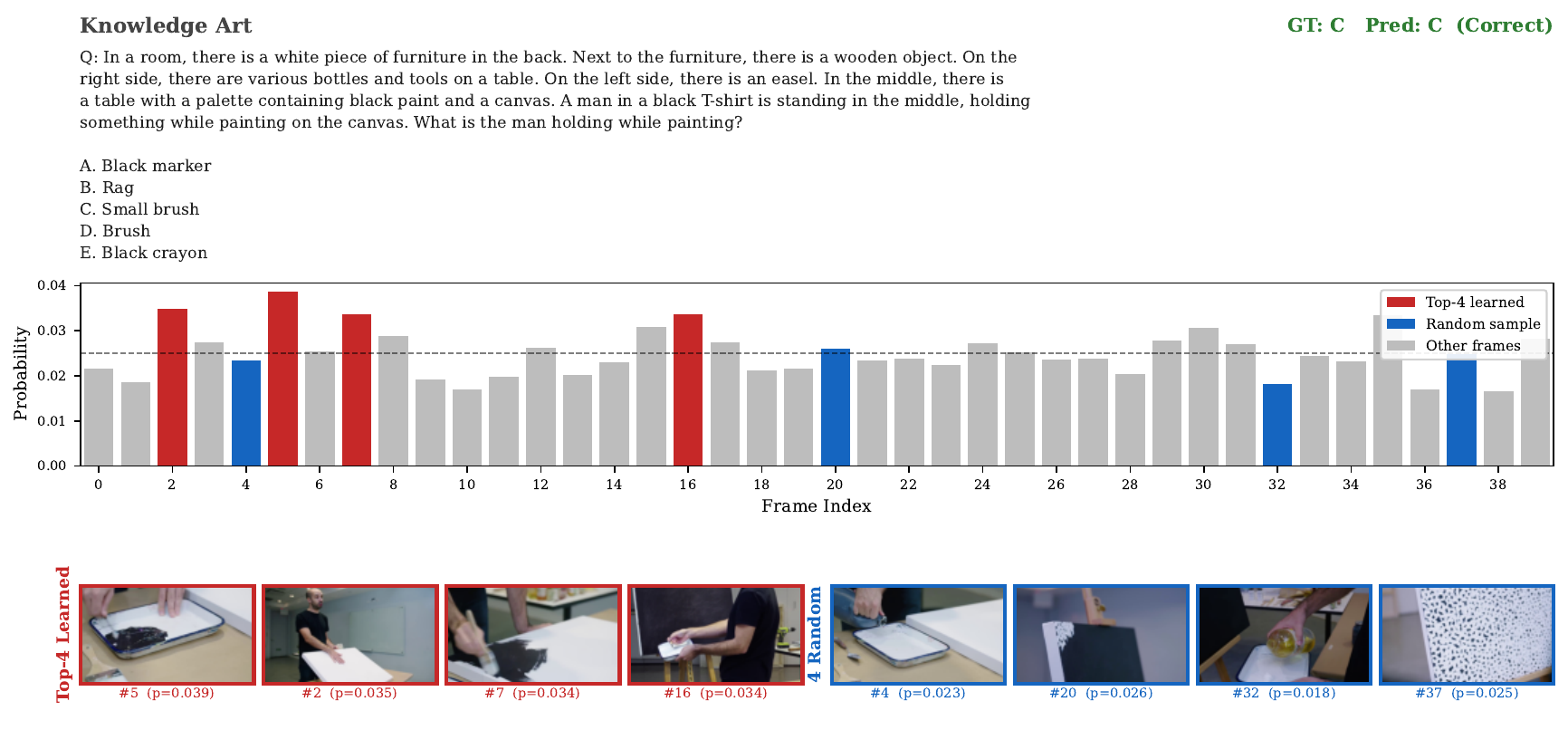}
        \caption{\textbf{Correct Prediction.} It can be observed that the third frame from learned distribution is important to correctly answer the question, as the person in this frame is holding on to a brush.}
        \label{fig:correct1}
    \end{subfigure}
    \begin{subfigure}[t]{\linewidth}
        \centering
        \includegraphics[width=\linewidth]{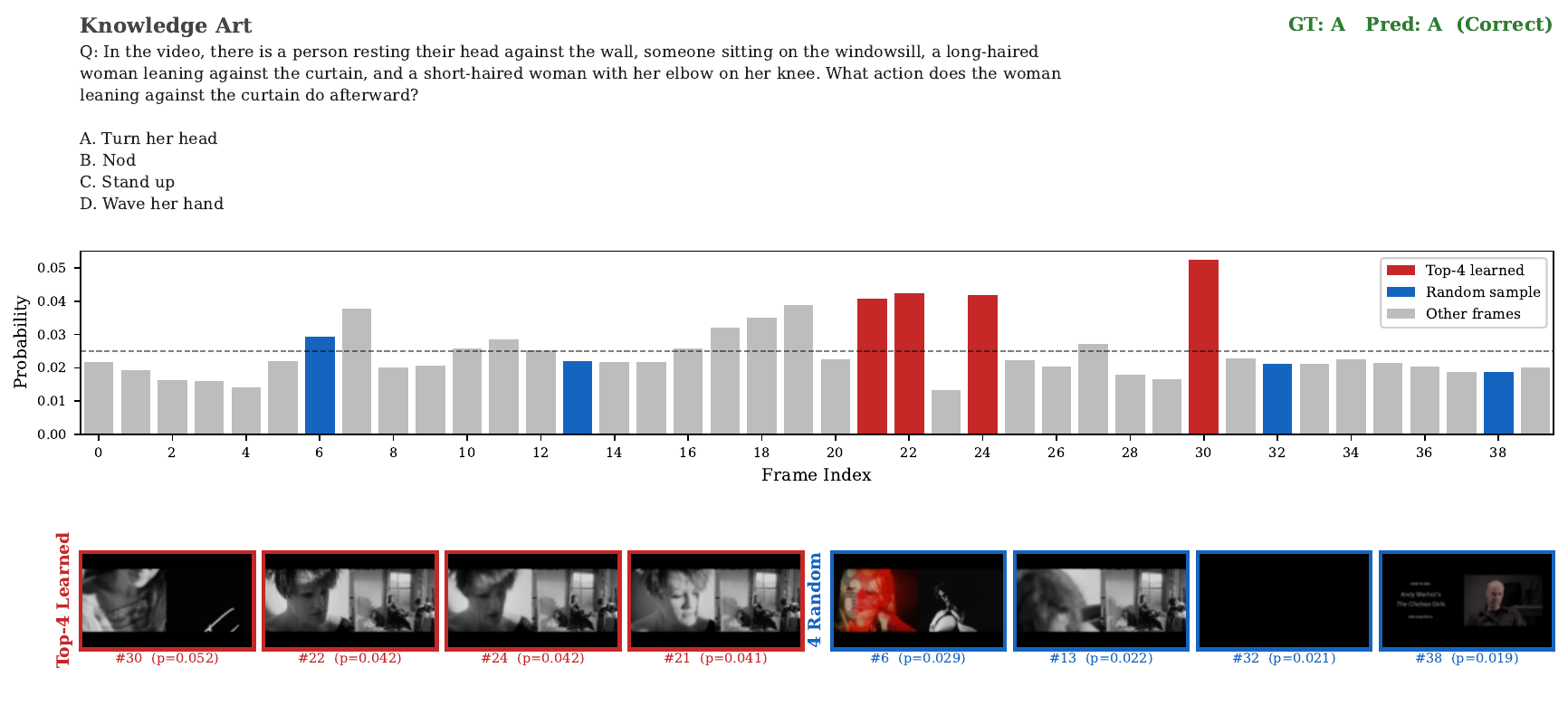}
        \caption{\textbf{Correct Prediction.} Frames sampled from learned distribution contain the scene where women turns her head, leading to the correct answer prediction.}
        \label{fig:correct2}
    \end{subfigure}
    \begin{subfigure}[t]{\linewidth}
        \centering
        \includegraphics[width=\linewidth]{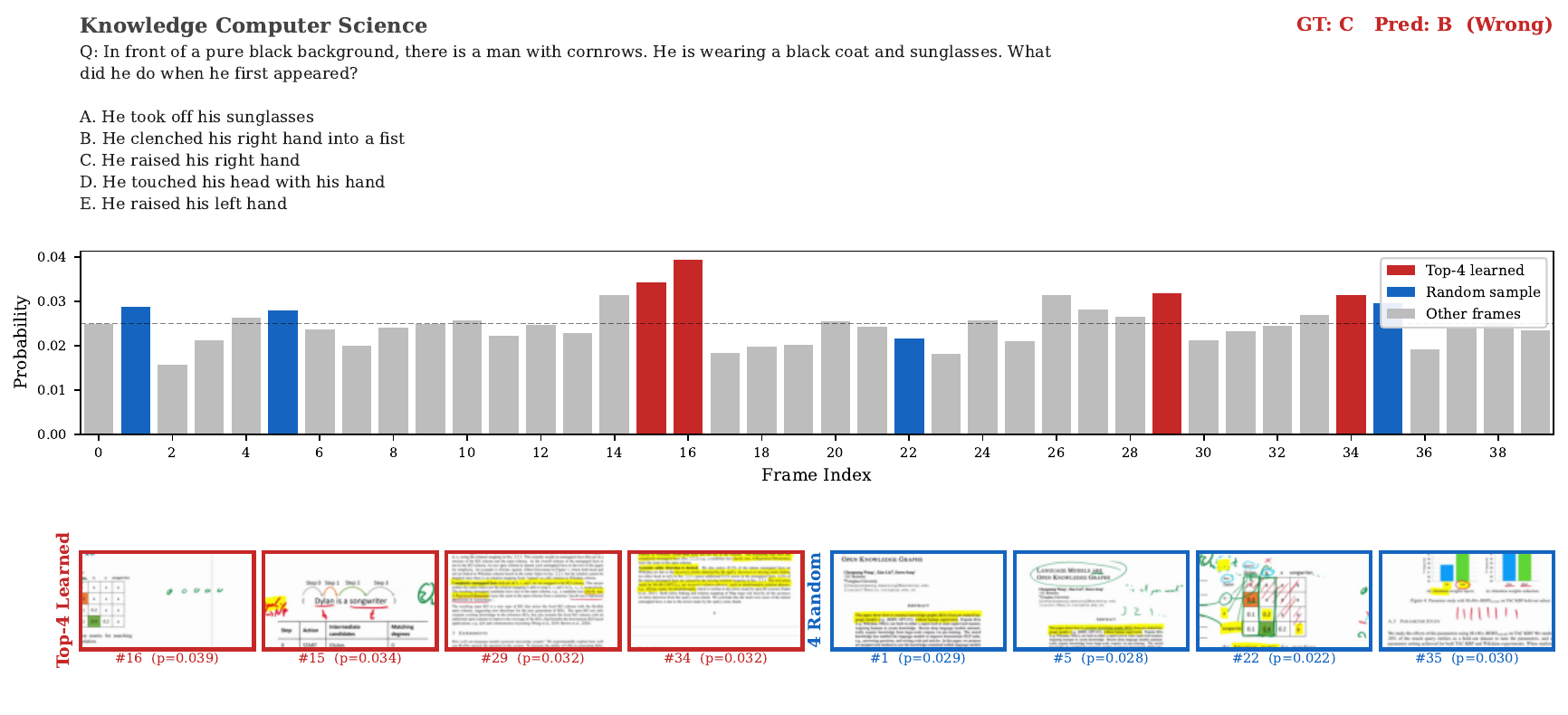}
        \caption{\textbf{Incorrect Prediction.} For this example, containing dense and cluttered visual information, TTA-Vid fails to select the most relevant frames, which can lead to incorrect predictions.}
        \label{fig:incorrect}
    \end{subfigure}

    \caption{\textbf{Qualitative Examples from LongVideoBench dataset.}}
    \label{fig:quals_lvb}
\end{figure*}

\section{Continual training across datasets}
\label{subsec:continual_training}
We consider a continual training setup to examine how effectively the model adapts as new datasets are introduced over time. This experiment evaluates the robustness and generalization ability of TTA-Vid under sequential domain shifts.
In Table~\ref{tab:continual_exp}, we report the results of the continual training setup. At each stage, the model is trained for only one epoch on the corresponding dataset. In Stage 1, the base Qwen2.5VL-7B model is trained using 32 samples from the MMVU dataset. In Stage 2, training continues on the VideoMMMU dataset, which consists of three splits; the model is sequentially trained on each split (perception → comprehension → adaptation), with one epoch per split. Subsequently, in Stage 3, the model is trained on the SciVideoBench dataset, and in Stage 4, it is trained on the Video-MME dataset. For evaluation, we test the model on the dataset corresponding to the most recent training stage (highlighted in blue in Table~\ref{tab:continual_exp}) and report the performance improvement relative to the base model. The results show that the proposed method effectively adapts to newly introduced datasets and consistently outperforms the base model under the continual training setup.

\section{Majority voting and multi-arm bandit (MAB)}
\label{subsec:motivation}
\paragraph{Why majority voting works.}
Our approach relies on majority voting across multiple rollouts from different frame views, to estimate a reference answer at test time, based on findings presented in prior test-time reinforcement learning approaches such as TTRL \cite{zuo2025ttrl} and TTRV \cite{singh2025ttrv}. From the observations made in TTRL \cite{zuo2025ttrl}, even though the majority answer may not always correspond to the true ground-truth label, it can still provide a useful signal for guiding learning during inference. The key intuition is that the verifier operates through answer comparison: when a predicted answer differs from the estimated reference answer, the model receives a negative reward. Even if the estimated label itself is incorrect, this comparison-based mechanism still often produces the correct directional feedback, since incorrect predictions are penalized while predictions consistent with the dominant answer are reinforced. Consequently, this process can provide stable optimization signals, allowing TTRL-style methods to remain effective even when the reference label obtained through majority voting is imperfect. In the case of videos, if the model reaches the same conclusion across $K$ different visual "views," the probability that it is a "hallucination" decreases, as hallucinations are typically less stable when the input signal changes.

\paragraph{Multi-arm Bandit (MAB).}
Frame selection in long videos requires both exploration of diverse frame subsets with exploitation of frames that contribute to higher-quality predictions. In our setting, multiple subsets of frames are sampled during each batch of rollouts, and the resulting answers are evaluated using the batched reward mechanism. Since rewards are already computed for each sampled subset as part of the test-time adaptation, these signals can be directly reused to update a frame-importance distribution. We formulate this update using a multi-arm bandit (MAB) framework, where each sampled subset acts as an arm and the observed reward reflects its usefulness for answer generation. The frame sampling distribution is updated using rewards obtained from batched rollouts. This requires no additional supervision, as the optimization naturally leverages the same reward signals used for model adaptation. Consequently, the model gradually prioritizes more informative frames while still maintaining exploration over alternative subsets, leading to improved frame selection compared to heuristic baselines (Table~\ref{tab:frame_selection_baseline}).

\section{Computational cost and test-time scaling (TTS)}
\label{subsec:runtime_analysis}
Methods such as Video-R1 \cite{feng2025video} and Video-RTS \cite{wang2025videorts} require large-scale training for both supervised fine-tuning (SFT) and reinforcement learning (RL). In contrast, TTA-Vid performs training once on a single batch of 32 samples and generalizes to all unseen QA pairs. On MMVU (625 QA pairs), TTA-Vid with InternVL3-2B as the base model completes both adaptation and evaluation in $12{,}097$\,s, achieving $56.99$\% accuracy. As a test-time scaling baseline, we compare against self-consistency. Self-consistency requires $18{,}617$\,s to achieve $55.04$\% accuracy, which is $35$\% slower while obtaining lower performance. Notably, for methods such as self-consistency, inference time scales directly with the number of test samples, whereas TTA-Vid adapts using only 32 samples and can then be evaluated directly on the entire test set.
To further analyze computational cost, we observe that runtime scales linearly with subsets and rollouts: doubling the number of subsets ($K{=}4 \rightarrow 8$) increases runtime by $2.1\times$ ($7{,}347$\,s $\rightarrow$ $15{,}481$\,s), while doubling rollouts ($N{=}8 \rightarrow 16$) increases runtime by $1.9\times$ ($7{,}347$\,s $\rightarrow$ $14{,}036$\,s). Importantly, the overall cost is independent of dataset size, as training is performed per batch (containing 32 samples).

\section{Qualitative analysis}
\label{subsec:quals}

Figure~\ref{fig:quals_lvb} presents qualitative examples from the LongVideoBench dataset. As illustrated, the learned distribution prioritizes frames containing task-relevant information compared to random sampling, leading to correct answer predictions for the former cases (see Figures~\ref{fig:correct1} and \ref{fig:correct2}). However, in particularly challenging video-question pairs where key visual cues are highly brief, the distribution may occasionally under-sample the critical frames, resulting in the incorrect prediction shown in the latter case (Figure~\ref{fig:incorrect}).

\end{document}